\pdfoutput=1

\documentclass[11pt]{article}

\usepackage[preprint]{acl}

\usepackage{comment} 

\usepackage{tabularx}

\usepackage{times}
\usepackage{latexsym}

\usepackage[T1]{fontenc}

\usepackage[utf8]{inputenc}

\usepackage{microtype}
\usepackage{enumitem}

\usepackage{inconsolata}

\usepackage{graphicx}

\usepackage[utf8]{inputenc}
\usepackage{booktabs} 
\usepackage{siunitx} 
\usepackage{adjustbox} 
\usepackage{multicol}
\usepackage{multirow}
\usepackage{adjustbox}
\usepackage{bbding}
\usepackage{amssymb}
\usepackage{siunitx}
\usepackage[most]{tcolorbox}
\usepackage{pifont}
\usepackage[table]{xcolor}
\usepackage{makecell}
\usepackage{xcolor}
\usepackage{textcomp}
\usepackage{fontawesome5}

\definecolor{LightCyan}{rgb}{0.88, 1, 1}
\definecolor{ForestGreen}{rgb}{0.13, 0.55, 0.13}

\newtcolorbox{blackpromptbox}[1][]{%
breakable,
colback=gray!10,
colframe=black,
fontupper=\small,
left=0.5mm, right=0.5mm, top=1mm, bottom=1mm,
boxrule=0.8pt,
sharp corners,
title={Prompt},
fonttitle=\bfseries,
#1 
}
\newtcolorbox{orangepromptbox}[1][]{%
    breakable,
    colback=orange!10!white,
    colframe=orange!70!black,
    fontupper=\small,
    left=0.5mm, right=0.5mm, top=1mm, bottom=1mm,
    boxrule=0.8pt,
    sharp corners,
    title={Prompt},
    fonttitle=\bfseries,
    #1
}

\newtcolorbox{bluepromptbox}[1][]{%
    breakable,
    colback=blue!5!white,
    colframe=blue!50!black,
    fontupper=\small,
    left=0.5mm, right=0.5mm, top=1mm, bottom=1mm,
    boxrule=0.8pt,
    sharp corners,
    title={Prompt},
    fonttitle=\bfseries,
    #1
}

\newtcolorbox{redpromptbox}[1][]{%
    breakable,
    colback=red!7!white,
    colframe=red!60!black,
    fontupper=\small,
    left=0.5mm, right=0.5mm, top=1mm, bottom=1mm,
    boxrule=0.8pt,
    sharp corners,
    title={Prompt},
    fonttitle=\bfseries,
    #1
}

\newtcolorbox{greenpromptbox}[1][]{%
    breakable,
    colback=green!8!white,
    colframe=green!60!black,
    fontupper=\small,
    left=0.5mm, right=0.5mm, top=1mm, bottom=1mm,
    boxrule=0.8pt,
    sharp corners,
    title={Prompt},
    fonttitle=\bfseries,
    #1
}

\newtcolorbox{lightorangepromptbox}[1][]{%
    breakable,
    colback=orange!5!white,
    colframe=orange!50!black,
    fontupper=\small,
    left=0.5mm, right=0.5mm, top=1mm, bottom=1mm,
    boxrule=0.8pt,
    sharp corners,
    title={Prompt},
    fonttitle=\bfseries,
    #1
}

\newtcolorbox{lightbluepromptbox}[1][]{%
    breakable,
    colback=cyan!6!white,
    colframe=blue!30!black,
    fontupper=\small,
    left=0.5mm, right=0.5mm, top=1mm, bottom=1mm,
    boxrule=0.8pt,
    sharp corners,
    title={Prompt},
    fonttitle=\bfseries,
    #1
}

\newtcolorbox{purplepromptbox}[1][]{%
    breakable,
    colback=violet!7!white,
    colframe=purple!60!black,
    fontupper=\small,
    left=0.5mm, right=0.5mm, top=1mm, bottom=1mm,
    boxrule=0.8pt,
    sharp corners,
    title={Prompt},
    fonttitle=\bfseries,
    #1
}

\title{\raisebox{-4pt}{\includegraphics[width=24pt]{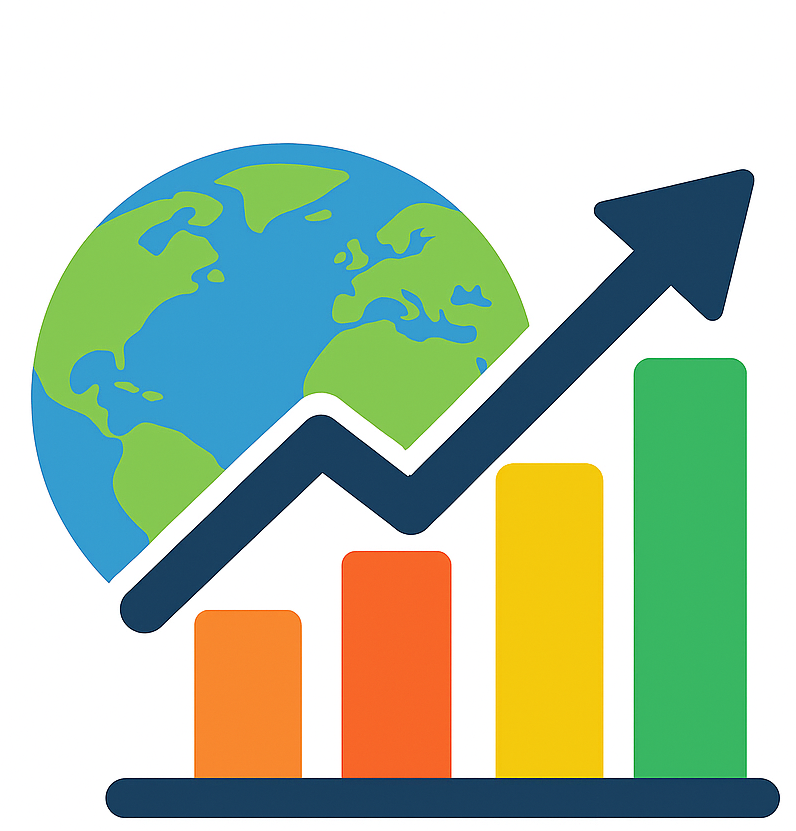}}\textsc{ PolyChartQA}: Benchmarking Large Vision-Language Models with\\ Multilingual Chart Question Answering}

\author{Yichen Xu$^{1}$\thanks{~~Equal Contribution.}
~~~Liangyu Chen$^{1*}$ ~~~Liang Zhang$^1$\\ 
\bf Jianzhe Ma$^1$
~~~Wenxuan Wang$^{1}$\thanks{~~Corresponding authors.}
~~~Qin Jin$^{1\dagger}$ \\ 
$^1$Renmin University of China \\
\texttt{\{xu\_yichen, liangyuchen, zhangliang00, majianzhe, wangwenxuan, qjin\}@ruc.edu.cn}
}

\begin{document}
\maketitle


\begin{abstract}
Charts are a universally adopted medium for data communication, yet existing chart understanding benchmarks are overwhelmingly English-centric, limiting their accessibility and relevance to global audiences.
To address this limitation, we introduce \textbf{\textsc{PolyChartQA}}, the first large-scale multilingual benchmark for chart question answering, comprising 22,606 charts and 26,151 QA pairs across 10 diverse languages. \textsc{PolyChartQA} is constructed through a scalable pipeline that enables efficient multilingual chart generation via data translation and code reuse, supported by LLM-based translation and rigorous quality control.
We systematically evaluate multilingual chart understanding with \textsc{PolyChartQA} on state-of-the-art LVLMs and reveal a significant performance gap between English and other languages, particularly low-resource ones.
Additionally, we introduce a companion multilingual chart question answering training set, \textsc{PolyChartQA}-Train, on which fine-tuning LVLMs yields substantial gains in multilingual chart understanding across diverse model sizes and architectures. Together, our benchmark provides a foundation for developing globally inclusive vision-language models capable of understanding charts across diverse linguistic contexts.
\end{abstract}

\section{Introduction}

Charts are ubiquitous tools for visualizing quantitative data and supporting analytical reasoning across domains such as science, business, and journalism, making accurate chart interpretation essential for data-driven decision-making. Recent advances in large vision-language models (LVLMs) have enabled significant progress in perceiving and reasoning over visualizations such as plots, diagrams, and charts. These models have shown promising results on tasks including complex chart question answering~\cite{masry2022chartqa, xia2024chartx, wang2024charxiv, masry2025chartqaprodiversechallengingbenchmark}, chart summarization~\cite{rahman2022chartsumm, tang2023vistext}, and chart image re-generation~\cite{moured2024chartformer, yang2024chartmimic}.

\begin{figure}[t]
    \includegraphics[width=\columnwidth]{images/Main_intro.png}
    \caption{
    Example of inconsistent chart understanding by LVLMs. The model answers correctly in English but fails on the Hindi equivalent.
    }
    \vspace{-10pt}
    \label{fig:intro}
\end{figure}

\begin{figure*}[!htbp] 
  \centering 
  \includegraphics[width=\textwidth]{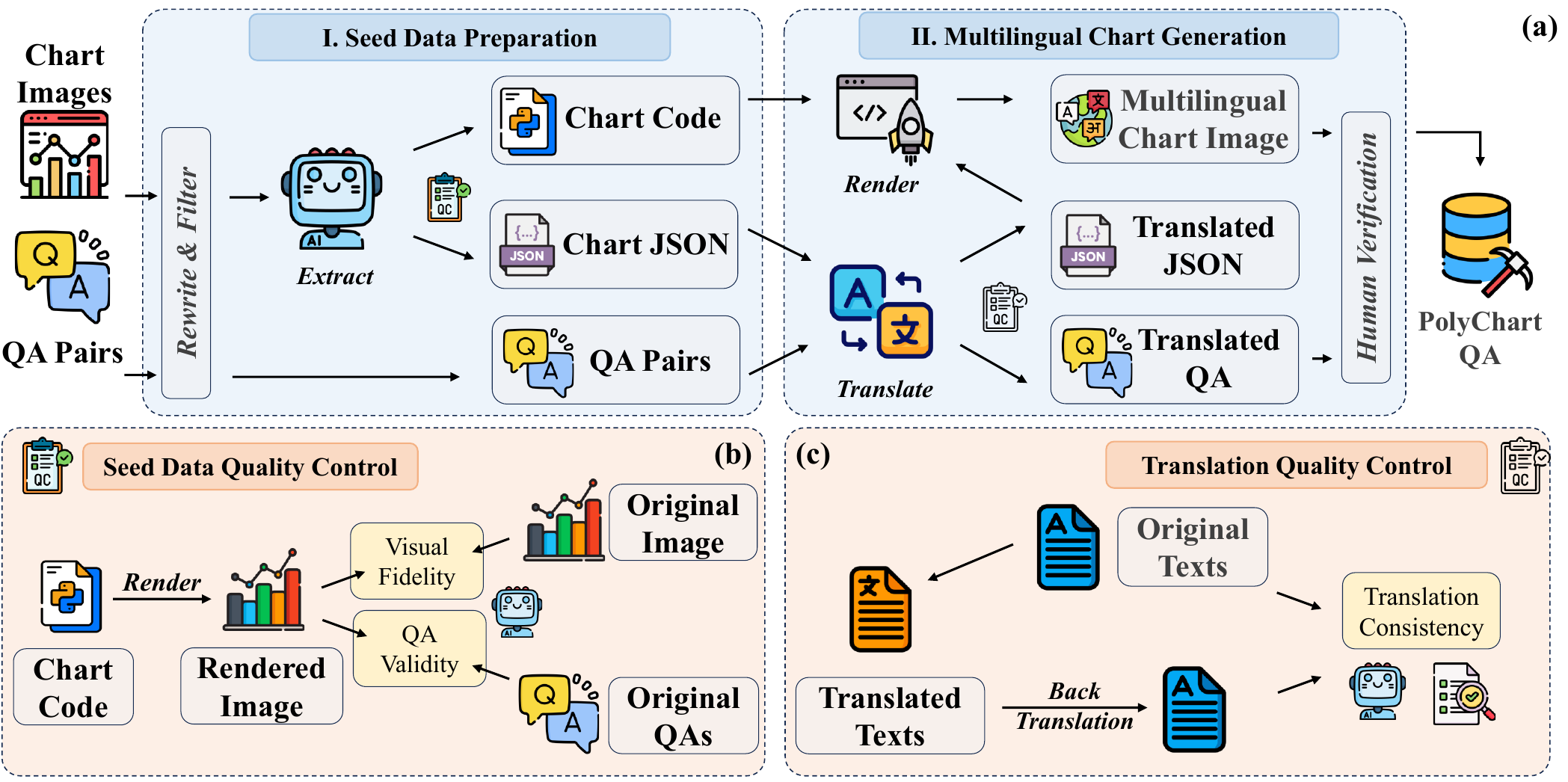} 
  \caption{Overview of the \textsc{PolyChartQA} data pipeline. (a) The full workflow consists of two stages: \textbf{Seed Data Preparation} and \textbf{Multilingual Chart Generation}. (b) Quality control procedures applied with seed data generation. (c) Quality control procedures applied during the translation stage.}
  \vspace{-10pt}
  \label{fig:pipeline} 
\end{figure*}

However, existing benchmarks for chart understanding remain overwhelmingly English-centric, overlooking the unique challenges of multilingual comprehension. As shown in Figure~\ref{fig:intro}, leading LVLMs often succeed on English chart QA but struggle with their non-English versions. This English-dominant bias poses a major barrier to developing globally inclusive chart understanding models, especially for underrepresented languages. While recent works~\cite{chen2024onechart, heakl2025kitab} have introduced bilingual chart datasets, they remain limited in scale and language coverage. To date, \textbf{no} comprehensive benchmark exists for evaluating multilingual chart understanding in LVLMs. Moreover, most multilingual multimodal benchmarks~\cite{pfeiffer2021xgqa, liu2021visually, yu2025cross, liu2024mmbench, xuan2025mmlu} focus on natural images rather than structured data like charts, leaving multilingual chart understanding largely unexplored. A key reason for this gap is the high cost of multilingual chart annotation~\cite{romero2024cvqa,tang2024mtvqa}, which severely restricts the scalability of such benchmarks.

To overcome these challenges, we develop \textsc{PolyChartQA} through a scalable two-stage pipeline. In the first stage, we generate high-quality English seed data by decomposing charts into structured JSON specifications and reusable code templates. In the second stage, we employ state-of-the-art LLMs to translate chart data and QA pairs and automatically render multilingual charts. A dedicated multi-stage quality-control procedure, combining automated consistency checks with final human verification, ensures the accuracy and naturalness of the multilingual data. Using this pipeline, we construct \textbf{\textsc{PolyChartQA}}, the first large-scale benchmark for multilingual chart understanding, spanning 10 widely spoken languages, including English, Chinese, Hindi, Spanish, French, Arabic, Bengali, Russian, Urdu, and Japanese, which together account for over 65\% of the global population~\citep{maaz2024palo}. The benchmark comprises a test set of over 22K chart images with 26K QA pairs and a training set of 751K QA pairs across 131K charts, providing a diverse and rigorously curated resource for evaluating and advancing multilingual chart understanding.

Using \textsc{PolyChartQA}, we present the first systematic evaluation of multilingual chart question answering in LVLMs, revealing that (i) current models remain markedly weak on multilingual chart QA, especially for low-resource languages, and (ii) cross-lingual generalization is fragile, with large performance gaps across scripts and sensitivity to partial visual–textual alignment. To bridge this gap and enhance multilingual chart capabilities, we show that fine-tuning on \textsc{PolyChartQA}-Train across different model families yields substantial performance gains, highlighting the effectiveness of instruction tuning for multilingual chart reasoning. We further provide a detailed error analysis across languages, scripts, and question types to expose persistent failure modes. In summary, our main contributions are:
\begin{itemize}[leftmargin=*, itemsep=0.2em]
    \item \textbf{Unified multilingual chart construction pipeline.} We propose a reproducible automatic pipeline for constructing high-quality, large-scale multilingual chart QA datasets.
    \item \textbf{\textsc{PolyChartQA} benchmark.} We introduce \textsc{PolyChartQA}, the first benchmark enabling systematic evaluation of LVLMs on chart understanding in ten diverse languages.
    \item \textbf{Comprehensive empirical analysis.} We conduct extensive experiments and error analysis that reveal critical performance gaps and demonstrate how our datasets substantially narrow them.
\end{itemize}

\begin{figure*}[tbp]
  \centering
  \includegraphics[width=0.99\textwidth]{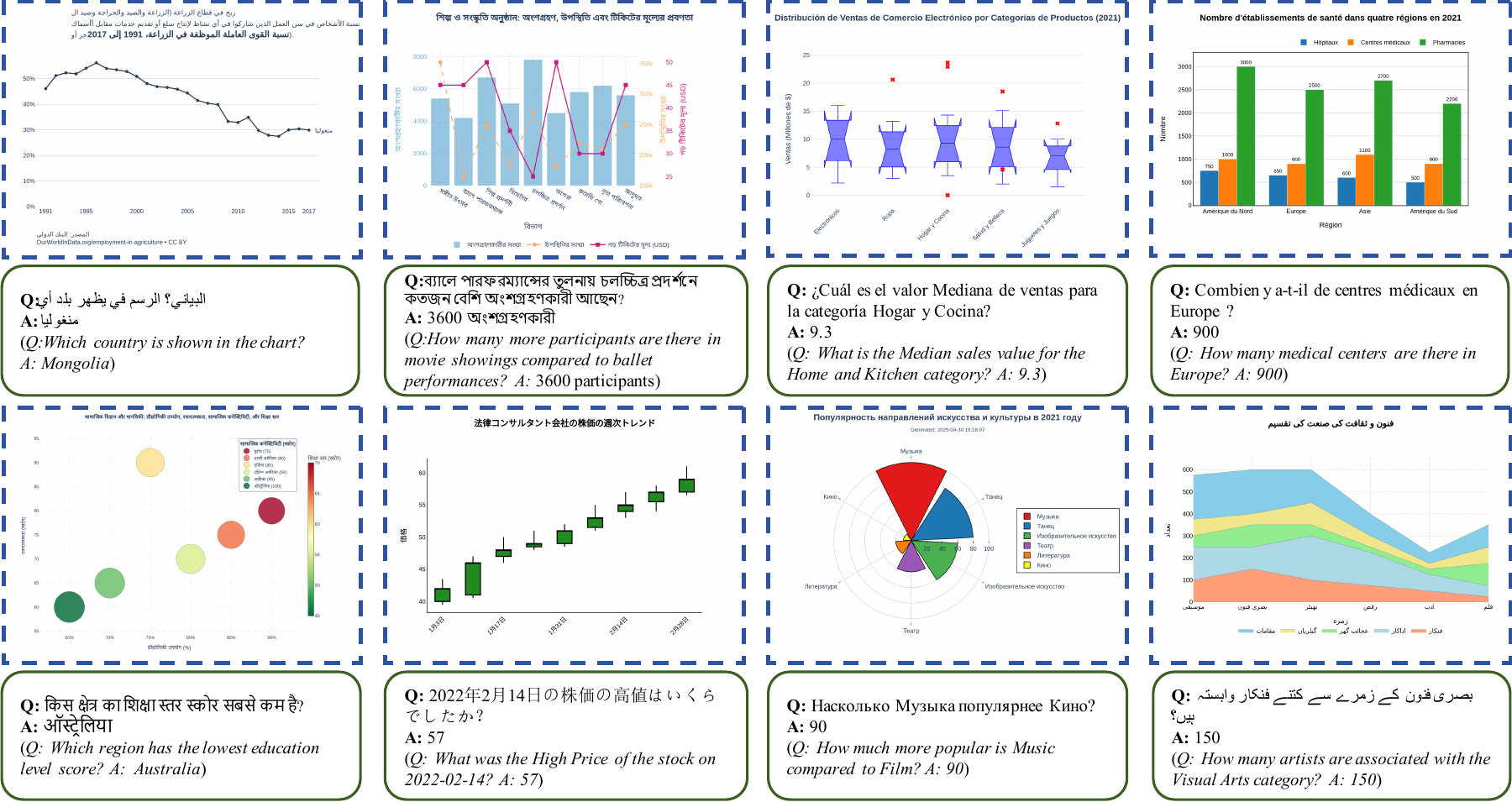}
  \caption{
    Multilingual chart question answering visualizations selected from \textsc{PolyChartQA}.
    First row, from left to right: Arabic, Bengali, Spanish, French. Second row, from left to right: Hindi, Japanese, Russian, Urdu.
  }
  \vspace{-10pt}
  \label{fig:case_study}
\end{figure*}

\section{Related Work}

\subsection{Chart Understanding Datasets}

Chart understanding requires models to jointly reason over visual and textual cues under diverse instructions. Recent benchmarks evaluate LVLMs on chart question answering~\cite{masry2022chartqa, methani2020plotqa, kantharaj2022opencqa}, summarization~\cite{tang2023vistext, kantharaj2022chart, rahman2022chartsumm}, chart-to-table conversion~\cite{xia2023structchart, xia2024chartx, chen2024onechart}, and re-rendering~\cite{moured2024chartformer, yang2024chartmimic}, with QA serving as the primary measure of fine-grained comprehension. Early datasets \cite{kahou2017figureqa,kafle2018dvqa,methani2020plotqa} mainly used synthetic charts and template-based questions, limiting diversity and realism. Later benchmarks~\cite{masry2022chartqa,xia2024chartx,liu2023mmc} moved toward realistic charts and human-authored questions, improving chart coverage and question complexity. However, most benchmarks remain English-only~\cite{chen2024onechart,heakl2025kitab}, limiting comprehensive evaluation and real-world deployment of LVLMs.

\subsection{Multilingual LVLMs}

Building on foundational monolingual models~\cite{li2023blip, team2024gemma,team2024gemma_2}, numerous multilingual LVLMs have emerged. Early influential works~\cite{chen2022pali,geigle2023mblip,beyer2024paligemma, steiner2024paligemma} pioneered scalable multilingual vision-language alignment. More recent open-source efforts such as PALO~\cite{maaz2024palo}, Maya~\cite{alam2024mayainstructionfinetunedmultilingual}, Pangea~\cite{yue2024pangea}, and Centurio~\cite{geigle2025centurio}, together with model families including QwenVL~\cite{Qwen-VL, bai2025qwen2, wang2024qwen2}, InternVL~\cite{chen2024expanding, chen2024far, chen2024internvl}, and Phi-Vision~\cite{abdin2024phi,abdin2024phi_4}, further broaden language coverage and improve multilingual multimodal performance. However, their ability to handle complex, text-rich visuals such as multilingual charts remains underexplored.

\subsection{Multilingual Evaluations on LVLMs}

The rapid progress of multilingual LVLMs has led to numerous benchmarks evaluating their multimodal capabilities, including general cross-lingual VQA~\cite{pfeiffer2021xgqa, changpinyo2022maxm}, text-centric VQA~\cite{tang2024mtvqa, yu2025cross}, and culturally grounded VQA~\cite{romero2024cvqa, liu2021visually, vayani2024all}. Comprehensive suites such as MMBench~\cite{liu2024mmbench}, MMLU-Prox~\cite{xuan2025mmlu}, and M4U~\cite{wang2024m4u} further assess reasoning, dialogue, captioning, and math problem solving, while M3Exam~\cite{zhang2023m3exam} and Exams-V~\cite{das2024exams} provide large-scale multilingual evaluations.
However, chart-based understanding remains largely underexplored, with limited coverage in existing benchmarks~\cite{zhang2023m3exam, geigle2025centurio}.

\section{\textsc{PolyChartQA}}



We present \textbf{\textsc{PolyChartQA}}, a large-scale multilingual chart question answering benchmark that addresses the scarcity of multilingual resources for chart understanding. As summarized in Table~\ref{tab:dataset_comparison}, \textsc{PolyChartQA} spans 10 languages (English, Chinese, Hindi, Spanish, French, Arabic, Bengali, Russian, Urdu, and Japanese) and covers 16 diverse chart types. The dataset is built through a unified pipeline (Figure~\ref{fig:pipeline}): we first construct high-quality English seed data comprising chart images, rendering code, structured JSON, and QA pairs, and then expand it to other languages via an LLM-assisted translation pipeline. The decoupled code-and-JSON representation further supports easy extension to related chart tasks (e.g., summarization and chart generation) without additional manual annotation. To ensure accuracy and reliability, we apply multi-stage quality control that combines automated validation with targeted human review. The remainder of this section details seed data preparation (§\ref{sec:seed}), multilingual chart generation (§\ref{sec:multi}), and quality control (§\ref{sec:quality}); additional pipeline details and prompts are provided in Appendix~\ref{app:pipeline_detail} and Appendix~\ref{app:full_prompts}, respectively.

\subsection{Seed Data Preparation}
\label{sec:seed}

We ground multilingual generation in high-quality English chart QA data by selecting three widely used benchmarks—ChartQA~\cite{masry2022chartqa}, ChartLlama~\cite{han2023chartllama}, and ChartX~\cite{xia2024chartx}—for their chart diversity, question coverage, and data quality. We construct \textsc{PolyChartQA}-Test from the test splits of ChartQA and ChartX, and \textsc{PolyChartQA}-Train from the training splits of ChartQA and ChartLlama; detailed statistics are summarized in Table~\ref{tab:chart_dataset_comparison}.

To ensure the quality of the seed data, we apply a two-step cleaning and validation procedure. \textbf{(i) Answer verification.} We use \textit{Gemini-2.5-Pro} to automatically check each chart question–answer pair; if the model’s prediction disagrees with the ground truth but suggests a clear correction, we manually revise the answer, otherwise we discard the sample. \textbf{(ii) Answer standardization.} We normalize verbose answers into concise canonical forms while preserving their semantics (e.g., ``the highest bar value in the chart is 42.1'' $\rightarrow$ ``42.1''). A manual review of 10\% of the cleaned data yields a pass rate above 98\%, confirming the reliability of the seed datasets.

Subsequently, we adopt a decoupled chart representation that separates content from visual rendering~\cite{shinoda2024sbs}, enabling flexible multilingual generation: the same rendering code can be reused with translated JSON to produce chart images in different languages. For each cleaned chart instance, we prompt \textit{Gemini-2.5-Pro} to generate two complementary artifacts: (i) a structured \textbf{JSON file} encoding the underlying data table, chart type, colors, and layout attributes, and (ii) an executable \textbf{Python script} that reproduces the chart using \textit{Plotly}~\footnote{\url{https://github.com/plotly/plotly.py}}, which natively supports multilingual text rendering.

\begin{table}[tbp]
    \small
    \centering
    \renewcommand{\arraystretch}{1.2}
    \resizebox{\columnwidth}{!}{ 
        \begin{tabular}{l c c r r}
            \toprule
            \textbf{Dataset} & \textbf{\#Lang.} & \makecell[c]{\textbf{Chart} \\ \textbf{Types}} & \textbf{\#Charts} & \textbf{\#QAs} \\
            \midrule
            ChartQA~\citeyearpar{masry2022chartqa} & 1 & 3 & 1,612 & 2,500 \\
            ChartX~\citeyearpar{xia2024chartx} & 1 & 18 & 1,152 & 2,304 \\
            ChartY~\citeyearpar{chen2024onechart} & 2 & 4 & 6,000 & 6,000 \\
            KITAB-Bench~\citeyearpar{heakl2025kitab} & 1 & 16 & 576 & 576 \\
            SMPQA~\citeyearpar{geigle2025centurio} & 11 & 2 & 1,100 & 4,300 \\
            ChartMind~\citeyearpar{wei2025chartmind} & 2 & 7 & 757 & 757 \\
            \midrule
            \textbf{PolyChartQA} & \textbf{10} & \textbf{16} & \textbf{22,606} & \textbf{26,151} \\
            \bottomrule
        \end{tabular}
    }
    \caption{Comparison of different chart-related datasets and benchmarks.}
    \label{tab:dataset_comparison}
\end{table}

\subsection{Multilingual Chart Generation}
\label{sec:multi}

To construct multilingual chart QA datasets, we translate the English seed data into multiple target languages via a two-stage process. We first obtain multilingual textual annotations (JSONs and QA pairs), and then render the corresponding chart images in each target language by reusing the template code.

\paragraph{Text Translation.}
Standard machine translation systems often struggle to preserve the structure and fine-grained semantics of chart-oriented JSON files and their associated QA pairs. In contrast, recent work~\cite{qiu2022multilingual, chen2023breaking, maaz2024palo} has shown that LLM-based translation achieves higher fidelity and consistency. Building on this, we adopt an LLM-based workflow with \textit{Gemini-2.5-Pro}, which jointly translates each chart’s JSON data and QA pairs to ensure semantic coherence. The model is instructed to preserve meaning while adapting to cultural and linguistic conventions to reduce translation bias. Our analyses in §\ref{sec:quality} indicate that the resulting multilingual corpora largely preserve the semantic content and structural properties of the original English data.

\paragraph{Chart Image Translation.}
Given the translated JSONs and QA pairs, we generate multilingual chart images by pairing each translated JSON with its corresponding template code and rendering the chart in the target language.

\subsection{Quality Control}
\label{sec:quality}

Our pipeline incorporates a multi-stage quality control mechanism to ensure both the accuracy and usability of the constructed dataset across all languages.

\paragraph{Seed Data Quality Control.}
To ensure the integrity of the English seed dataset, we applied a multi-stage validation process, as shown in Figure~\ref{fig:pipeline}~(b). With both JSON files and rendering code acquired, we first executed the code to verify reproducibility and automatically removed any samples that failed to render successfully. We then examined two key aspects of data quality. \textbf{(i) Visual Fidelity}: Each regenerated chart was compared against its original version using \textit{Gemini-2.5-Pro} to detect visual or semantic discrepancies. Charts showing notable mismatches in chart type, data values, or layout were discarded. \textbf{(ii) QA Validity}: We further verified that all questions remained answerable from the reconstructed charts, using \textit{Gemini-2.5-Pro} and \textit{GPT-4.1} as independent validators. Both models possess strong vision–language reasoning and code understanding capabilities, and requiring agreement between them provides a stricter and more reliable validation process. Only samples confirmed as valid by both models were retained, removing those with semantic inconsistency or linguistic errors.

\begin{figure}[t]
  \includegraphics[width=\columnwidth]{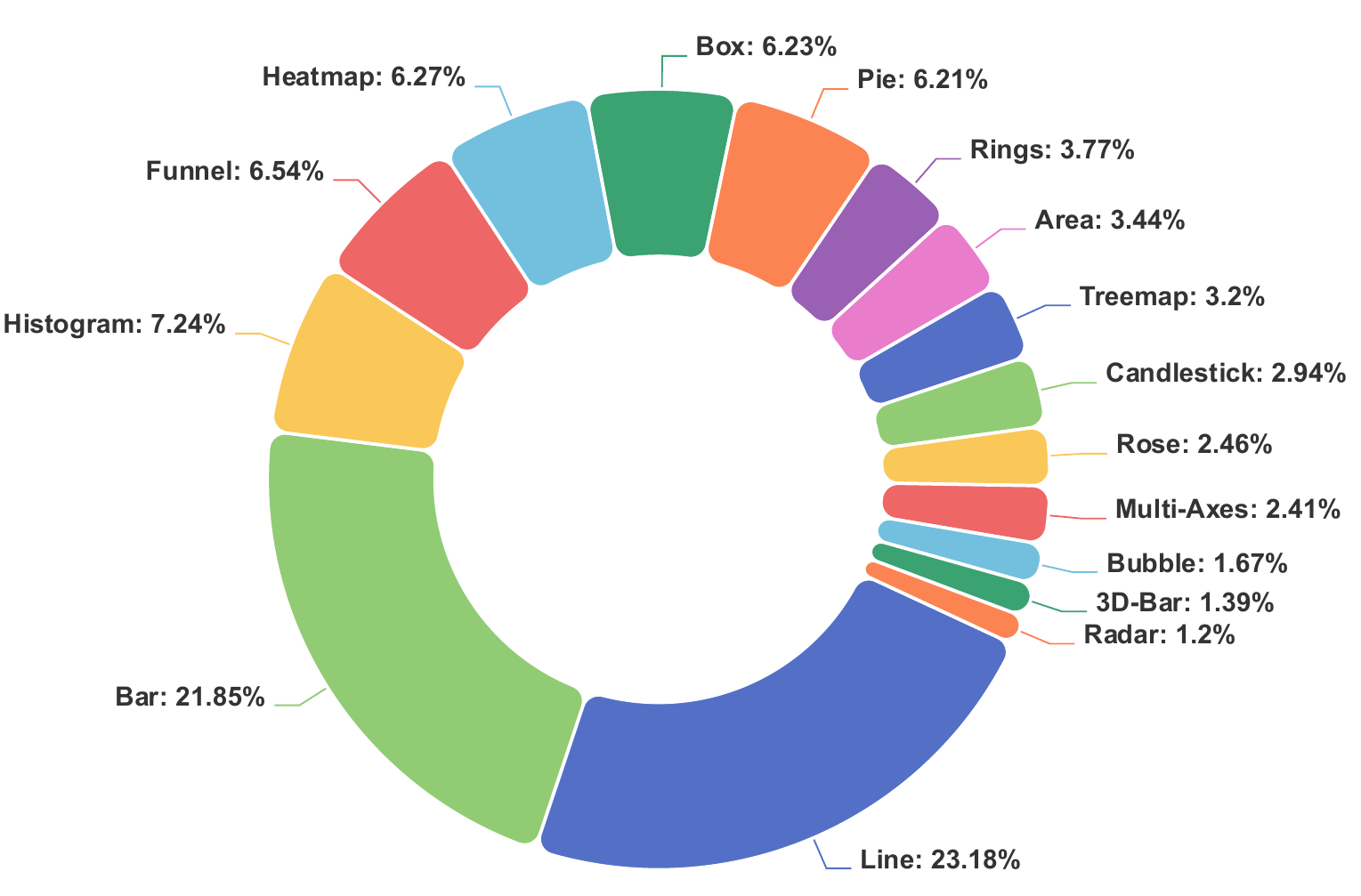}
  \caption{
    Distribution of chart types in \textsc{PolyChartQA}.
  }
  \vspace{-5pt}
  \label{fig:chart_types}
\end{figure}

\paragraph{Multilingual Data Quality Control.}
Building upon the validated seed data, we further applied a two-stage quality control procedure to ensure the reliability of the multilingual outputs.
Similar to the seed stage, any samples whose chart code failed to execute during the multilingual image generation stage were automatically discarded. For the remaining data, we evaluated both text translation quality and multilingual chart image quality. \textbf{(i) Translation Quality}: As illustrated in Figure~\ref{fig:pipeline}(c), each translated instance was back-translated into English and compared with the original. We assessed textual consistency using the METEOR~\cite{banerjee2005meteor} metric, complemented by semantic judgements from \textit{Gemini-2.5-Pro} to compensate for METEOR’s limited sensitivity to nuanced meaning differences. Samples with back-translated content that deviated substantially from the original English semantics were filtered out. \textbf{(ii) Visual Inspection}: All remaining multilingual chart images were then manually reviewed to identify and remove those containing visual defects such as text clipping, misaligned layouts, or rendering artifacts.

\begin{table}[!tbp]
\small
\centering
\setlength{\tabcolsep}{3pt} 
    \begin{tabular}{@{}lccc@{}}
    \toprule
    \textbf{Metrics} & {\textbf{\begin{tabular}{@{}c@{}}Image\\Quality\end{tabular}}} & {\textbf{\begin{tabular} {@{}c@{}}QA\\Relevance\end{tabular}}} & {\textbf{\begin{tabular}{@{}c@{}}Translation\\Accuracy\end{tabular}}} \\
    \midrule
    \textbf{Avg. Score}   & 2.87 & 2.93 & 2.89 \\
    \textbf{Avg. Disag.}       & 3.1  & 3.4  & 4.1  \\
    \textbf{Avg. $\bar{\kappa}_w$}   & 0.887 & 0.817 & 0.885 \\
    \bottomrule
    \end{tabular}%
\vspace{-5pt}
\caption{Average human scores and inter-annotator agreement scores for each evaluation dimension. "Disag." shows the raw count of differing ratings and $\kappa_w$ denotes weighted Cohen’s $\kappa$.}
\label{tab:human_eval_summary}
\end{table}

\begin{table*}[tbp]
\small
\centering
\begin{adjustbox}{max width=\textwidth}
\begin{tabular}{@{}lc ccccccccccccc@{}}
\toprule
\textbf{Model} & \textbf{\#Params} & {\textbf{EN}} & {\textbf{ZH}} & {\textbf{FR}} & {\textbf{ES}} & {\textbf{RU}} & {\textbf{JA}} & {\textbf{AR}} & {\textbf{UR}} & {\textbf{HI}} & {\textbf{BN}} & {\textbf{\begin{tabular}{@{}c@{}}Avg.\\(w/ EN)\end{tabular}}} & {\textbf{\begin{tabular}{@{}c@{}}Avg.\\(w/o EN)\end{tabular}}} \\
\midrule
\multicolumn{14}{@{}l}{~~\textbf{\textit{Proprietary Models}}} \\
GPT-4o & - & \underline{55.9} & \underline{46.0} & \underline{53.4} & \underline{54.4} & \underline{52.4} & \underline{45.4} & \underline{50.5} & \underline{48.7} & \underline{51.3} & \underline{48.2} & \underline{50.9} & \underline{50.2} \\
Gemini-2.5-Pro & - & \textbf{70.6} & \textbf{67.7} & \textbf{69.0} & \textbf{69.3} & \textbf{67.6} & \textbf{68.6} & \textbf{69.1} & \textbf{67.5} & \textbf{68.6} & \textbf{66.0} & \textbf{68.5} & \textbf{68.2} \\
\midrule
\multicolumn{14}{@{}l}{~~\textbf{\textit{Open Source Models}}} \\
InternVL-2.5~\citep{chen2024expanding} & 2B & 27.8 & 3.3 & 14.7 & 9.2 & 9.5 & 2.0 & 4.3 & 0.3 & 1.2 & 0.1 & 7.8 & 5.1 \\
InternVL-3~\citep{zhu2025internvl3} & 2B & 43.7 & 35.3 & 30.8 & 33.5 & 25.6 & 26.9 & 17.1 & 14.6 & 15.7 & 11.9 & 25.6 & 23.1 \\
Qwen2-VL~\citep{wang2024qwen2} & 2B & 42.3 & 33.6 & 37.6 & 37.7 & 35.9 & 22.2 & 28.8 & 19.1 & 24.4 & 23.0 & 30.7 & 29.1 \\
Qwen2.5-VL~\citep{bai2025qwen2} & 3B & \textbf{67.4} & \textbf{59.6} & \textbf{61.8} & \textbf{62.5} & \textbf{58.0} & \underline{48.8} & \underline{51.4} & \underline{37.2} & \underline{45.7} & \underline{43.0} & \underline{53.7} & \underline{51.8} \\
PaliGemma2~\citep{steiner2024paligemma} & 3B & 26.6 & 14.7 & 19.7 & 21.5 & 13.9 & 10.7 & 15.9 & 12.2 & 14.3 & 10.2 & 16.3 & 14.9 \\
Phi-3.5-Vision~\citep{abdin2024phi} & 4.2B & 45.1 & 17.5 & 37.2 & 36.9 & 26.9 & 15.7 & 9.3 & 4.7 & 10.6 & 10.6 & 23.2 & 20.2 \\
DeepSeek-VL2~\citep{wu2024deepseekvl2mixtureofexpertsvisionlanguagemodels} & 4.5B & 40.1 & 38.8 & 26.4 & 34.1 & 19.9 & 0.0 & 14.2 & 13.8 & 19.1 & 16.3 & 24.8 & 22.5 \\
Phi-4 Vision~\citep{abdin2024phi_4} & 5.6B & \underline{62.3} & 46.0 & 55.9 & 44.6 & 48.7 & 41.6 & 29.7 & 23.4 & 33.4 & 18.3 & 40.6 & 37.7 \\
LLaVA-OneVision~\citep{llava_one_vision} & 7B & 18.7 & 10.1 & 13.1 & 14.2 & 9.4 & 8.3 & 7.5 & 5.2 & 7.1 & 5.7 & 10.1 & 9.0 \\
LLaVA-v1.6~\citep{llava1_6} & 7B & 24.8 & 12.9 & 18.9 & 18.2 & 13.5 & 11.5 & 12.0 & 7.7 & 10.0 & 6.7 & 13.9 & 12.4 \\
Qwen2-VL~\citep{wang2024qwen2} & 7B & 56.4 & 54.3 & 53.4 & 52.7 & 52.2 & 47.3 & 40.5 & 32.0 & 43.9 & 40.3 & 47.3 & 46.1 \\
Qwen2.5-VL~\citep{bai2025qwen2} & 7B & 60.5 & \underline{58.3} & \underline{57.2} & \underline{59.0} & \underline{56.8} & \textbf{55.6} & \textbf{52.0} & \textbf{43.7} & \textbf{49.4} & \textbf{46.4} & \textbf{53.8} & \textbf{53.0} \\
InternVL-2.5~\citep{chen2024expanding} & 8B & 39.2 & 26.3 & 32.4 & 33.5 & 29.5 & 22.6 & 10.9 & 11.2 & 14.0 & 13.4 & 23.5 & 21.4 \\
InternVL-3~\citep{zhu2025internvl3} & 8B & 54.1 & 39.4 & 43.4 & 45.8 & 38.1 & 39.7 & 21.4 & 17.2 & 20.2 & 17.5 & 33.8 & 31.0 \\
Llama-3.2-Vision~\citep{llama_3_2_vision} & 11B & 15.5 & 16.9 & 14.1 & 12.9 & 15.4 & 9.6 & 13.1 & 14.4 & 21.3 & 17.5 & 15.2 & 15.2 \\
\midrule
\multicolumn{14}{@{}l}{~~\textbf{\textit{Chart Specific Models}}} \\

TinyChart~\citep{zhang2024tinychart} & 3B & \textbf{45.6} & 15.1 & 23.5 & \underline{26.7} & 12.3 & 11.1 & 10.7 & 9.3 & 10.6 & 7.9 & \textbf{17.9} & \underline{14.2} \\
ChartGemma~\citep{masry2025chartgemma} & 3B & 14.4 & 7.2 & 17.2 & \textbf{30.2} & 15.2 & 9.0 & 9.5 & 6.0 & \textbf{13.5} & 6.2 & 11.1 & 10.6 \\
ChartInstruct~\citep{masry2024chartinstruct} & 7B & 23.8 & \underline{15.2} & 21.2 & 21.7 & 16.6 & \underline{12.6} & 6.7 & 0.1 & 3.9 & 0.0 & 12.3 & 10.7 \\
ChartLlama~\citep{han2023chartllama} & 13B & 11.7 & 7.9 & \textbf{26.7} & 21.9 & \textbf{21.4} & 12.0 & \underline{11.8} & \textbf{15.6} & 10.6 & \textbf{13.1} & 15.6 & 14.1 \\
ChartAssistant~\citep{meng2024chartassisstant} & 13B & \underline{25.8} & \textbf{15.8} & \underline{25.1} & 24.4 & \underline{18.5} & \textbf{14.2} & \textbf{11.9} & \underline{11.7} & \underline{11.5} & \underline{9.3} & \underline{17.1} & \textbf{15.9} \\

\midrule
\multicolumn{14}{@{}l}{~~\textbf{\textit{Multilingual Models}}} \\
Centurio~\cite{geigle2025centurio} & - & 7.9 & 4.0 & 3.6 & 3.0 & 1.5 & 2.5 & 2.0 & 1.5 & 1.5 & 1.0 & 2.9 & 2.2 \\
Pangea~\cite{maaz2024palo} & 7B & \textbf{24.7} & \textbf{13.6} & \textbf{19.8} & \textbf{21.3} & \textbf{15.8} & \textbf{11.5} & \textbf{13.1} & \textbf{12.1} & \textbf{13.1} & \textbf{13.1} & \textbf{16.1} & \textbf{14.9} \\
PALO~\cite{maaz2024palo} & 7B & \underline{11.5} & 6.0 & \underline{10.5} & \underline{9.9} & \underline{7.0} & 5.9 & 7.0 & 5.0 & 5.2 & 3.6 & \underline{7.3} & \underline{6.7} \\
Maya~\cite{alam2024mayainstructionfinetunedmultilingual} & 8B & 8.7 & \underline{6.4} & 7.6 & 7.2 & 6.8 & \underline{6.0} & \underline{7.1} & \underline{5.7} & \underline{6.9} & \underline{5.6} & 6.8 & 6.6 \\
\bottomrule
\end{tabular}
\end{adjustbox}
\vspace{-8pt}
\caption{Overall performance on \textsc{PolyChartQA}. Bold values in each model category denote the best performance and underlined values denote the second best.}
\label{tab:model_accuracy_final_layout}
\end{table*}

\subsection{Data Statistics}
\label{sec:statistics}

\textsc{PolyChartQA} consists of 154,121 chart images and 777,514 question answer pairs across 10 languages, split into a test set (\textsc{PolyChartQA}-Test) with 22,606 charts and 26,151 QA pairs and a training set (\textsc{PolyChartQA}-Train) with 131,515 charts and 751,363 QA pairs. It spans 16 diverse chart types (Figure~\ref{fig:chart_types}), with representative examples shown in Figure~\ref{fig:case_study}. More detailed statistics of \textsc{PolyChartQA} are provided in Appendix~\ref{app:dataset_detail_statistics}.

To assess the quality of \textsc{PolyChartQA}-Test, we conduct a \textbf{human evaluation} on a randomly sampled 20\% subset for each language. Bilingual annotators rate each instance along three dimensions:
\textbf{(i) Translation Quality}, assessing semantic accuracy, fluency, and naturalness while avoiding bias or misinformation;
\textbf{(ii) Chart Image Quality}, evaluating visual clarity, text legibility, and overall presentation; and
\textbf{(iii) QA Correctness}, verifying question relevance and factual consistency with the chart. Each instance was annotated by one annotator and independently reviewed by another to ensure reliability.
As summarized in Table~\ref{tab:human_eval_summary}, all three dimensions achieve near-ceiling performance, with average scores above 2.8 (out of 3) and strong inter-annotator agreement ($\bar{\kappa}_w > 0.8$), confirming the overall reliability of \textsc{PolyChartQA}-Test. Additional details are provided in Appendix~\ref{app:human_eval}.

\section{Experiments}
\label{sec:experiments}

\subsection{Experimental Setup}
\label{sec:models}

To thoroughly assess the multilingual perception and reasoning abilities of modern LVLMs on our multilingual chart benchmark, we select 22 representative state-of-the-art models from four categories: open-source general MLLMs, open-source multilingual LVLMs, chart-specific LVLMs, and closed-source LVLMs. 

All baseline models are evaluated under their official configurations. During inference, we set the decoding temperature to $0.01$ and $top\_p$ to $0.7$. We use a unified multilingual prompt: "Answer the question using a word or phrase in <target\_language> or a number in digits. <Question>" All results are averaged over 8 independent runs. Experiments are conducted on 8 NVIDIA A100 GPUs.

\subsection{Evaluation Results}
\paragraph {Metrics.}
Following prior work~\cite{masry2022chartqa}, we adopt a type-aware relaxed accuracy metric: numerical predictions are considered correct if within 5\% relative error of the ground truth; non-numerical answers require exact string match.

\paragraph{Zero-shot Evaluation.}
\label{sec:zero-shot-results}




Table~\ref{tab:model_accuracy_final_layout} reports the zero-shot performance of various models on \textsc{PolyChartQA}. A substantial gap is observed between closed-source and open-source models: \textit{Gemini-2.5-Pro} achieves the best overall performance across all languages (Avg. 68.5), while \textit{GPT-4o} is notably lower (Avg. 50.9).

Among open-source models, \textit{Qwen2.5-VL} is the strongest, performing well across both high- and low-resource languages and even surpassing \textit{GPT-4o} on average. By comparison, \textit{InternVL-3} and \textit{DeepSeek-VL2} show larger drops on non-English inputs, indicating limited robustness for multilingual chart understanding.

Chart-specific models also struggle in multilingual settings, as prior chart-focused models that perform well in English fail to generalize effectively to other languages.
Multilingual LVLMs such as \textit{Pangea}, \textit{PALO}, \textit{Maya}, and \textit{Centurio} exhibit weak overall accuracy on \textsc{PolyChartQA}, suggesting that broad multilingual pretraining alone is insufficient for text-rich chart reasoning and grounding.

Across model families, accuracy is relatively stable for high-resource languages such as English, Chinese, and French, but degrades sharply for low-resource languages, particularly Urdu and Hindi, consistent with prior findings~\cite{maaz2024palo}. This trend indicates that current multilingual training pipelines provide insufficient chart-specific grounding in low-resource settings, likely due to data scarcity and imbalanced language representation.

\paragraph{Cross-lingual Performance Varies by Model Families.}
\label{sec:cross-lingual-results}


We evaluate four representative model families, Qwen2.5-VL, InternVL3, PaliGemma2, and LLaVA-v1.6, under cross-lingual input settings where either the chart image or the QA pair is replaced with its English counterpart, as shown in Table~\ref{tab:cross_lingual_condition_col}. We observe clear family-level differences as the linguistic alignment between modalities varies. Qwen2.5-VL achieves its best performance under fully aligned multilingual inputs, while introducing English into either modality slightly degrades accuracy, consistent with its strong zero-shot performance on non-English data and reliance on language-consistent visual–text alignment. In contrast, InternVL3, PaliGemma2, and LLaVA-v1.6 show improved accuracy when English is introduced, reflecting a heavier dependence on English as a pivot language to compensate for weaker non-English grounding. These results indicate that robust multilingual chart understanding requires exposure to diverse cross-lingual alignment patterns beyond English-centric supervision.

\begin{table}[h]
\small
\centering
\setlength{\tabcolsep}{5pt} 
\begin{tabularx}{\linewidth}{@{}c c c >{\centering\arraybackslash}X >{\centering\arraybackslash}X@{}}
\toprule
\textbf{\begin{tabular}{@{}c@{}}Model\\Size\end{tabular}} & \textbf{\begin{tabular}{@{}c@{}}Multi.\\Img.\end{tabular}} & \textbf{\begin{tabular}{@{}c@{}}Multi.\\QA\end{tabular}} &{\textbf{\begin{tabular}{@{}c@{}}Avg.\\(w/ EN)\end{tabular}}} & {\textbf{\begin{tabular}{@{}c@{}}Avg.\\(w/o EN)\end{tabular}}} \\
\midrule
\multirow{3}{*}{\textbf{\begin{tabular}{@{}c@{}}Qwen2.5-\\VL-3B\end{tabular}}}
 & \ding{55} & \checkmark & 49.6 & 47.3 \\
 & \checkmark & \ding{55} & 52.1 & 49.9 \\
 & \checkmark & \checkmark & \textbf{53.7} & \textbf{51.8} \\
\midrule
\multirow{3}{*}{\textbf{\begin{tabular}{@{}c@{}}Qwen2.5-\\VL-7B\end{tabular}}}
 & \ding{55} & \checkmark & 48.3 & 46.6 \\
 & \checkmark & \ding{55} & 51.0 & 49.5 \\
 & \checkmark & \checkmark & \textbf{53.8} & \textbf{53.0} \\
\midrule
\multirow{3}{*}{\textbf{\begin{tabular}{@{}c@{}}InternVL3\\-2B\end{tabular}}}
 & \ding{55} & \checkmark & \textbf{27.9} & \textbf{25.8} \\
 & \checkmark & \ding{55} & 27.6 & 25.2 \\
 & \checkmark & \checkmark & 25.6 & 23.1 \\
\midrule
\multirow{3}{*}{\textbf{\begin{tabular}{@{}c@{}}InternVL3\\-8B\end{tabular}}}
 & \ding{55} & \checkmark & \textbf{42.0} & \textbf{40.2} \\
 & \checkmark & \ding{55} & 37.6 & 34.8 \\
 & \checkmark & \checkmark & 33.8 & 31.0 \\
\midrule
\multirow{3}{*}{\textbf{\begin{tabular}{@{}c@{}}PaliGemma2-\\3B\end{tabular}}}
 & \ding{55} & \checkmark & \textbf{29.0} & \textbf{28.4} \\
 & \checkmark & \ding{55} & 18.6 & 17.1 \\
 & \checkmark & \checkmark & 16.3 & 14.9 \\
\midrule
\multirow{3}{*}{\textbf{\begin{tabular}{@{}c@{}}LLaVA-\\v1.6-7B\end{tabular}}}
 & \ding{55} & \checkmark & \textbf{18.0} & \textbf{16.3} \\
 & \checkmark & \ding{55} & 17.3 & 16.2 \\
 & \checkmark & \checkmark & 13.9 & 12.4 \\
\bottomrule
\end{tabularx}
\vspace{-5pt}
\caption{Cross-lingual performance of different LVLMs. \textit{Multi. Img.} and \textit{Multi. QA} indicate whether the chart image or QA pair is multilingual. Bold numbers denote the best results for each model.}
\label{tab:cross_lingual_condition_col}
\end{table}

\paragraph{Fine-tuning Significantly Boosts Multilingual Chart Understanding.}
\label{sec:training-results}

Multilingual chart comprehension poses a significant challenge for LVLMs. To address this limitation, we investigate a straightforward yet highly effective strategy: fine-tuning these models on dedicated multilingual chart instruction data using \textsc{PolyChartQA}-test. For a comprehensive evaluation, we selected 6 representative LVLMs spanning various architectures and sizes: Qwen2.5-VL-3B, Qwen2.5-VL-7B, InternVL3-2B, InternVL3- 8B, PaliGemma2-3B, and LLaVA-v1.6-Mistral-7B. We applied LoRA~\cite{hu2022lora} training with a rank of $r=128$ and a learning rate of $1e^{-5}$; the vision encoder was kept frozen, and all models were trained for a single epoch.

\begin{table}[!htbp]
\small
\centering
\setlength{\tabcolsep}{5pt} 
\begin{tabularx}{\linewidth}{l | >{\centering\arraybackslash}X >{\centering\arraybackslash}X}
\toprule
\textbf{Model} & {\textbf{\begin{tabular}{@{}c@{}}Avg.\\(w/ EN)\end{tabular}}} & {\textbf{\begin{tabular}{@{}c@{}}Avg.\\(w/o EN)\end{tabular}}} \\
\midrule
\textbf{Qwen2.5-VL-3B} & 53.7 & 51.8 \\
\hspace{1.5em}\textit{+ fine-tuning} & \textbf{61.1} \textcolor{ForestGreen}{\small (+13.8\%)} & \textbf{60.2} \textcolor{ForestGreen}{\small (+16.2\%)} \\
\textbf{Qwen2.5-VL-7B} & 53.8 & 53.0 \\
\hspace{1.5em}\textit{+ fine-tuning} & \textbf{66.9} \textcolor{ForestGreen}{\small (+24.3\%)} & \textbf{66.1} \textcolor{ForestGreen}{\small (+24.7\%)} \\
\midrule
\textbf{InternVL-3-2B} & 25.6 & 23.1 \\
\hspace{1.5em}\textit{+ fine-tuning} & \textbf{33.3} \textcolor{ForestGreen}{\small (+30.1\%)} & \textbf{31.2} \textcolor{ForestGreen}{\small (+35.1\%)} \\
\textbf{InternVL-3-8B} & 33.8 & 31.0 \\
\hspace{1.5em}\textit{+ fine-tuning} & \textbf{44.0} \textcolor{ForestGreen}{\small (+30.2\%)} & \textbf{41.4} \textcolor{ForestGreen}{\small (+33.5\%)} \\
\midrule
\textbf{PaliGemma2-3B} & 16.3 & 14.9 \\
\hspace{1.5em}\textit{+ fine-tuning} & \textbf{29.0} \textcolor{ForestGreen}{\small (+77.9\%)} & \textbf{28.4} \textcolor{ForestGreen}{\small (+90.6\%)} \\
\midrule
\textbf{LLaVA-v1.6-7B} & 13.9 & 12.4 \\
\hspace{1.5em}\textit{+ fine-tuning} & \textbf{25.5} \textcolor{ForestGreen}{\small (+83.5\%)} & \textbf{24.0} \textcolor{ForestGreen}{\small (+93.5\%)} \\
\bottomrule
\end{tabularx}
\vspace{-5pt}
\caption{Fine-tuning results using \textsc{PolyChartQA}-Train across different model families and sizes. Performance gains are highlighted in green.}
\label{tab:training_results_sim}
\end{table}

\begin{figure*}[htbp]

  \includegraphics[width=\textwidth]{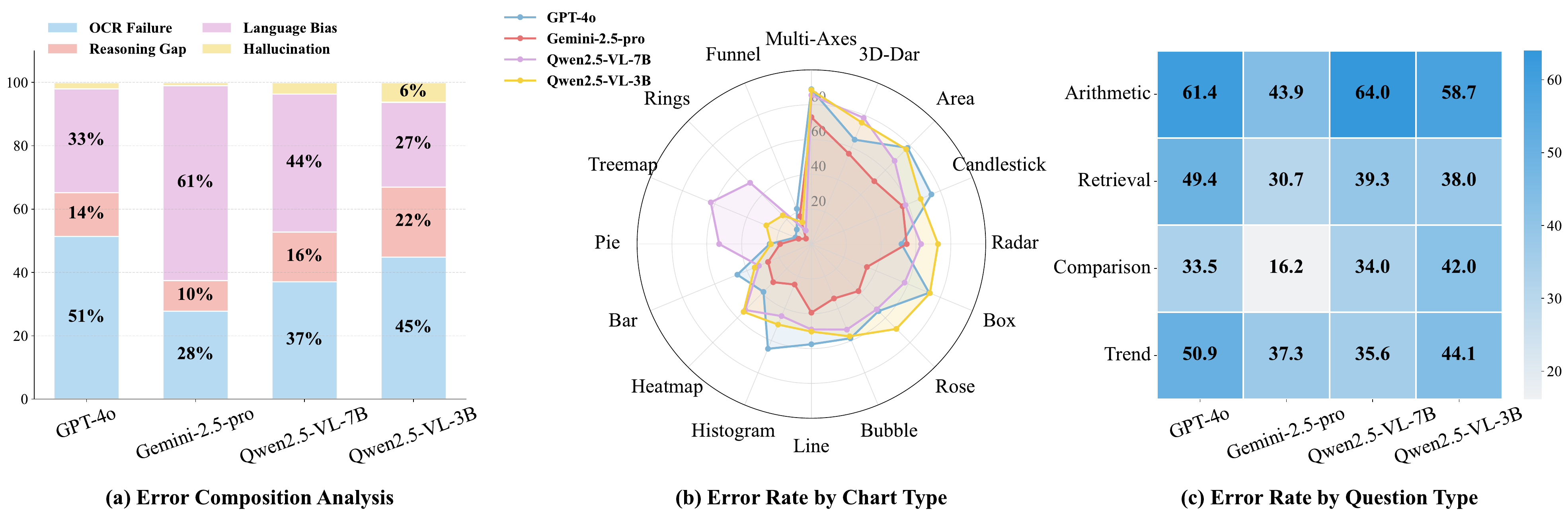}
  \caption{
   Error analysis across error types, chart types, and question types.
  }
  \vspace{-5pt}
  \label{fig:error_analysis}
\end{figure*}

\paragraph{Performance Scales with Training Data Size.}
\label{sec:training-ablation-data-percentage}

\begin{figure}[htbp]

  \includegraphics[width=\columnwidth]{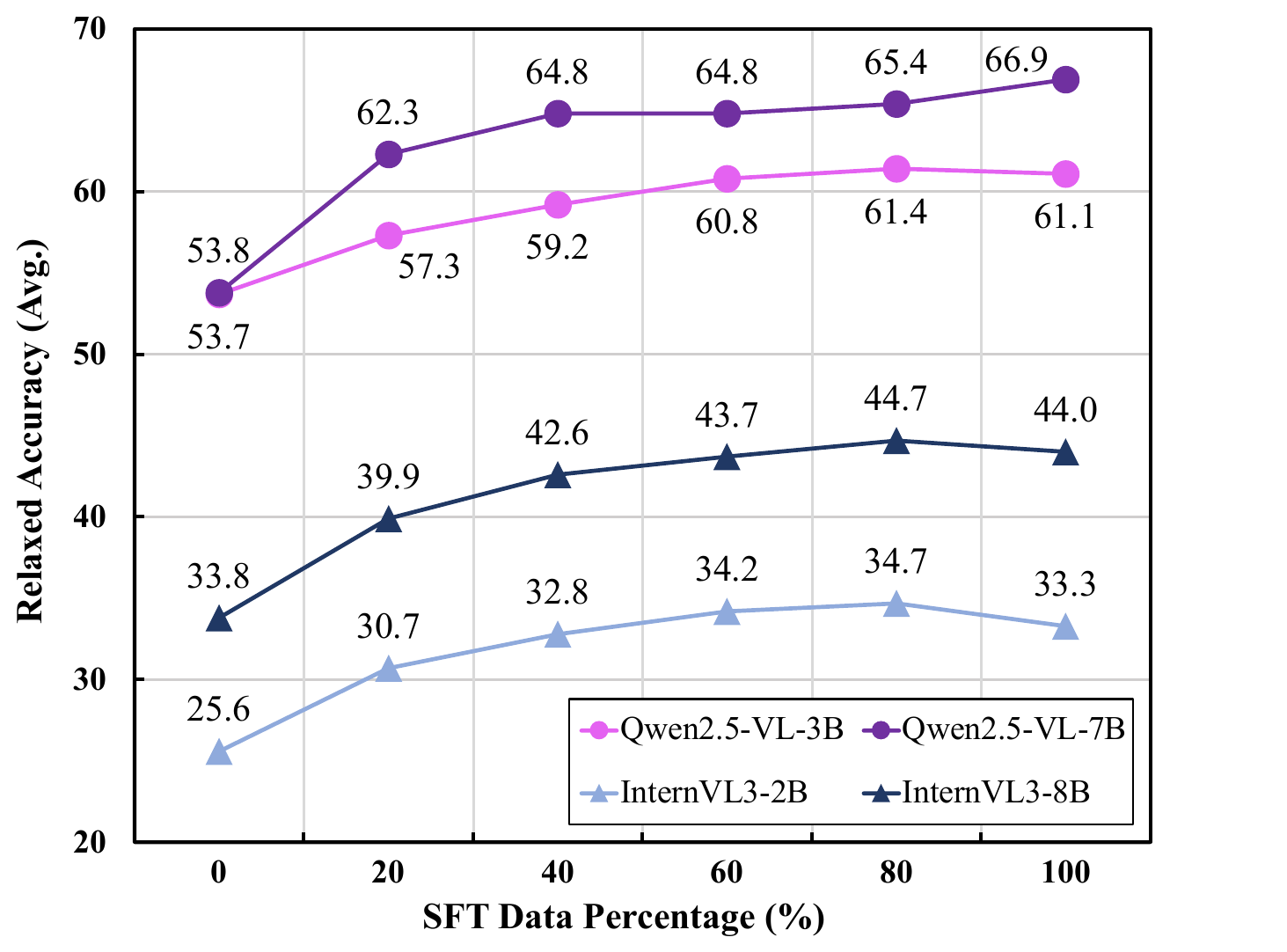}
  \caption{
   Performance on \textsc{PolyChartQA} with respect to the SFT data size across different model families.
  }
  \vspace{-5pt}
  \label{fig:ablation_data_percentage}
\end{figure}

As summarized in Table~\ref{tab:training_results_sim}, fine-tuning on \textsc{PolyChartQA}-Train yields \textbf{substantial performance improvements across all models}. The average accuracy increases by approximately 20\% for Qwen2.5-VL, 30\% for InternVL3, and over 70\% for PaliGemma2 and LLaVA-v1.6. Notably, Qwen2.5-VL-7B surpasses \textit{GPT-4o} and reaches performance comparable to \textit{Gemini-2.5-Pro} after fine-tuning. These results highlight \textbf{the strong generalizability and effectiveness of \textsc{PolyChartQA}} in enhancing multilingual chart understanding across diverse LVLM architectures.



We assess the impact of data scale by fine-tuning each model on 20\%–100\% of the \textsc{PolyChartQA}-Train. As shown in Figure~\ref{fig:ablation_data_percentage}, \textbf{performance scales positively with data size} across all model families and capacities. The most substantial gains occur within the initial 20\% of data, indicating that early exposure provides the greatest learning benefit~\cite{shaham2024multilingual}. Smaller models such as InternVL3-2B and Qwen2.5-VL 3B tend to reach performance saturation earlier, at around 80\%. Whereas stronger models such as Qwen2.5-VL-7B continue to benefit from additional data, demonstrating greater scalability and data utilization efficiency. These results suggest that larger models better capture diverse multilingual chart patterns, whereas smaller ones may benefit from more targeted or curriculum-based training.

\subsection{Error Analysis}

To further investigate model limitations, we selected four representative models and conducted a multi-dimensional error analysis. We first evaluated error rates across different chart and question types. To further diagnose the root causes of these errors, we sampled 300 failure cases per model for each language and categorized them manually. Overall and language-wise breakdowns exhibit similar trends across languages.

As shown in Figure~\ref{fig:error_analysis}(a), analysis reveals that OCR failures (27.8\%–51.4\%) and Language Bias (26.8\%–61.5\%) are the dominant error sources, together accounting for the vast majority of incorrect predictions. Reasoning gaps constitute a moderate portion (9.7\%–22.2\%), while Hallucination remains a minor issue (<7\%).
Figure~\ref{fig:error_analysis}(b) shows that error rates increase with chart complexity, with multi-axes, 3D-bar, and candlestick charts exhibiting substantially higher failure rates than simpler formats. Figure~\ref{fig:error_analysis}(c) further reveals clear variation across question types: arithmetic questions are the hardest, while comparison and retrieval are easier.


\section{Conclusion}

In this paper, we introduce \textbf{\textsc{PolyChartQA}}, the first large-scale multilingual benchmark for chart question answering, covering 10 diverse languages. Built through a scalable and reproducible pipeline, \textsc{PolyChartQA} enables efficient multilingual chart generation and evaluation. Experiments reveal that existing LVLMs struggle with multilingual chart understanding, particularly in non-Latin languages. Applying fine-tuning on \textsc{PolyChartQA}-Train leads to substantial and consistent improvements across all model architectures, demonstrating the effectiveness and strong generalizability of our dataset. We hope this work inspires broader research into multilingual multimodal understanding and foster the development of more inclusive, globally accessible LVLMs.

\section*{Limitations}

Despite introducing the first large-scale multilingual benchmark for chart question answering, \textsc{PolyChartQA} still has several limitations. While it includes a diverse set of major languages, it excludes many lesser-spoken or low-resource ones, limiting its global inclusivity. Secondly, since \textsc{PolyChartQA} builds on existing datasets, it may inherit framing biases or inaccuracies from the source datasets. Additionally, although we employ a multi-stage validation process with human review, the use of LLM-based generation and translation may still introduce subtle shifts in tone, cultural framing, or emphasis across languages. Future work may explore fully human-annotated datasets when feasible, extend \textsc{PolyChartQA} to additional chart understanding tasks beyond QA, and expand to more complex real-world visual formats such as infographics or interactive dashboards.

\section*{Ethics Statements}
Our work aims to promote language inclusivity and accessibility in AI technologies by constructing a multilingual benchmark focused on chart understanding. By systematically evaluating model performance across diverse languages and scripts, especially those underrepresented in existing resources, we highlight current limitations and foster the development of more equitable large vision-language models. We believe this contributes to reducing the dominance of English in AI systems and supports the global community in accessing AI tools in their native languages. We acknowledge that our dataset, being derived from existing sources, may inherit biases or misinformation from the original charts. Furthermore, our use of LLMs for translation, despite a multi-stage validation process, may introduce subtle artifacts such as tonal shifts or cultural inaccuracies. We encourage future work to further improve multilingual data fidelity and broaden the linguistic inclusivity of AI systems.

\bibliography{custom}

\appendix

\appendix

\section{Data Construction Pipeline Details}
\label{app:pipeline_detail}

This section provides extended technical details on our data construction pipeline, clarifying design choices, dataset selection, and quality assurance processes. It also addresses common concerns regarding technical contributions, source datasets selection, and filtering statistics.

\subsection{Source Dataset Selection}
To validate our choice of source datasets, Table~\ref{tab:chart_dataset_comparison} compares existing English chart QA datasets in terms of realism, diversity, and scale. We selected ChartQA and ChartX because they together provide an optimal combination of coverage, real-world grounding, and annotation quality, forming a strong foundation for multilingual extension.

\begin{table}[h]
\centering
\small
\setlength{\tabcolsep}{3pt} 
\begin{tabular}{lcccc}
\toprule

\textbf{Dataset} & \textbf{\begin{tabular}{@{}c@{}}Chart\\Types\end{tabular}} & \textbf{\begin{tabular}{@{}c@{}}Real-World\\Charts\end{tabular}} & \textbf{\#Charts} & \textbf{\#QAs} \\
\midrule
PlotQA & 3 & \ding{55} & 224K & 28M \\
ChartQA & 3 & \ding{51} & 21.9K & 32.7K \\
OpenCQA & 5 & \ding{51} & -- & -- \\
ChartBench & 9 & \ding{55} & 66.6K & 599.6K \\
ChartX & 18 & \ding{55} & 6K & 6K \\
\bottomrule
\end{tabular}
\caption{Comparison of major English chart QA datasets.}
\label{tab:chart_dataset_comparison}
\end{table}

ChartQA contributes high-quality, human-annotated real-world QA pairs, while ChartX adds diversity through synthetic chart types. Together, they balance realism, diversity, and usability, which is crucial for developing a representative multilingual benchmark.

\subsection{Source Dataset Licenses}

We use three existing chart QA datasets as part of our data construction pipeline. \textsc{ChartQA} is released under the GPL-3.0 license\footnote{\url{https://huggingface.co/datasets/ahmed-masry/ChartQA}}, \textsc{ChartX} under the CC-BY-4.0 license\footnote{\url{https://huggingface.co/datasets/U4R/ChartX}}, and \textsc{ChartLlama} under the MIT license\footnote{\url{https://huggingface.co/datasets/listen2you002/ChartLlama-Dataset}}. All datasets are publicly available via HuggingFace and used in accordance with their respective licenses.

\subsection{Language Definition}
We follow the language selection and the definition of high/low-resource languages in~\cite{maaz2024palo}, which identifies Arabic, Urdu, Hindi, and Bengali as low-resource languages among the ten included in our benchmark.

\subsection{Filtering Statistics and Data Retention}
We report detailed filtering ratios and retained item counts across all stages of data construction when constructing \textsc{PolyChartQA} to ensure transparency and reproducibility:

\begin{itemize}
    \item \textbf{Source Dataset Cleaning \& Validation:} 11.2\% filtered and 0.5\% corrected through automated validation; all items passed normalization (remaining: 7,545).
    \item \textbf{Seed Data Generation (with Quality Control):} 35.9\% filtered during JSON/code extraction and chart-type balancing (remaining: 4,840 core seed items).
    \item \textbf{Text Translation:} 23.2\% filtered across 10 languages after automated validation (remaining per language: $\sim$3,716).
    \item \textbf{Chart Image Translation:} 11.4\% removed after rendering validation (remaining total: 32,897).
    \item \textbf{Final Visual Inspection (in Multilingual Data Quality control):} 20.5\% filtered through manual inspection, resulting in a final dataset of 26,151 multilingual QA pairs.
\end{itemize}

These statistics demonstrate that each stage enforced strict quality thresholds, ensuring the reliability and linguistic–visual consistency of the final benchmark dataset. Since \textsc{\textsc{PolyChartQA}-Train} serves as the training set, we did not record detailed statistics for it.




\section{Detailed Dataset Statistics of \textsc{PolyChartQA}}
\label{app:dataset_detail_statistics}

This section provides detailed data statistics of \textsc{PolyChartQA}. It covers Data Statistics by Language and Chart Type, Question and Answer Length Statistics, Per-language Distribution of Images and Questions, as well as the Distribution of Images, Questions, JSON, and Code for the English seed data in \textsc{PolyChartQA}-Test (§~\ref{sec:appendix_polychartqa_test}), and Data Statistics by Language and Chart Type for \textsc{PolyChartQA}-Train (§~\ref{sec:appendix_polychartqa_train}).

\subsection{\textsc{PolyChartQA}-Test}
\label{sec:appendix_polychartqa_test}

\paragraph{Data Statistics by Language and Chart Type.}

We show the detailed statistics of \textsc{PolyChartQA} in Tables~\ref{tab:polychartqa_image_stats} and~\ref{tab:polychartqa_qa_stats}, including per-language and per-chart-type breakdowns for both images and QA pairs. Note that “EN” here does not refer to the original English dataset; instead, it was regenerated and processed through the same pipeline as other languages, with the only exception being the translation step.

\paragraph{Question and Answer Length Statistics.}

\begin{figure*}[htbp]
  \centering
  \includegraphics[width=\textwidth]{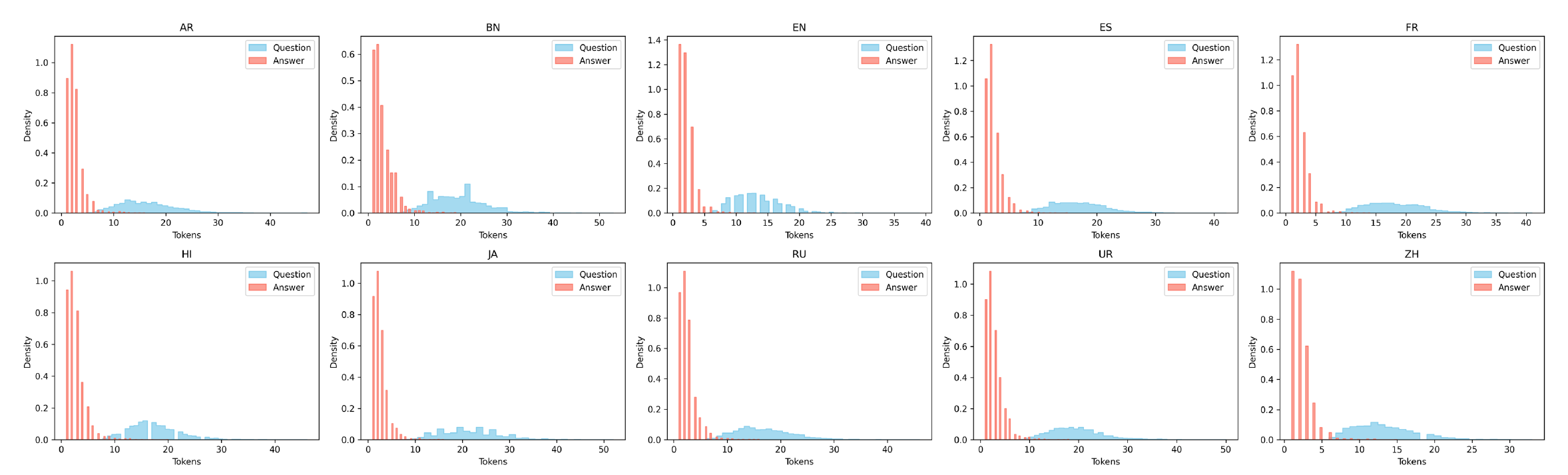}
  \caption{
    Question and answer length statistics in \textsc{PolyChartQA}.
  }
  \label{fig:qa_length_stats}
\end{figure*}

We report statistics of question and answer lengths across all ten languages in \textsc{PolyChartQA}, using token counts computed with the GPT-4o tokenizer. The distribution for each language, aggregated over training and test splits, is illustrated in Figure~\ref{fig:qa_length_stats}. These results highlight significant variation in textual length, which reflects both linguistic and orthographic diversity across languages.

\paragraph{Distribution of Images, Questions, JSON, and Code for English Seed Data.}

\begin{figure*}[htbp]
  \centering
  \includegraphics[width=\textwidth]{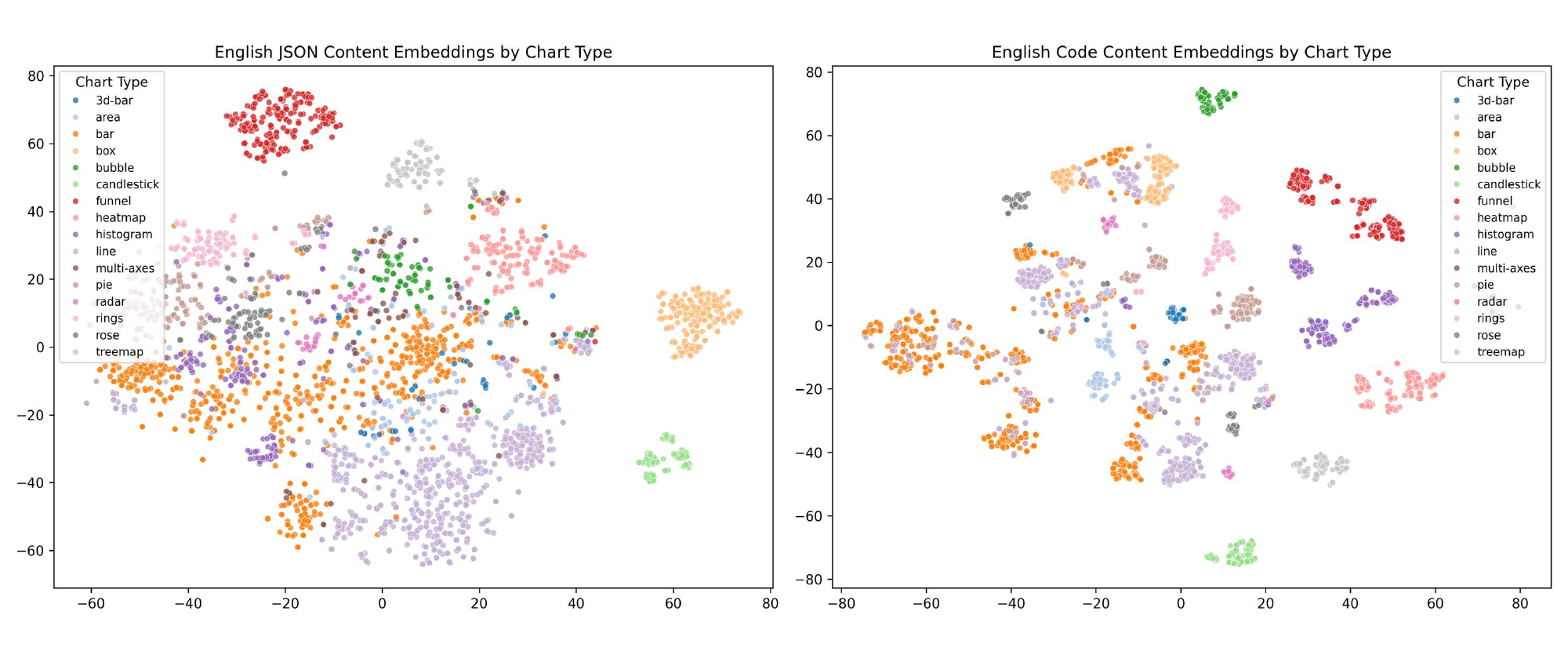}
  \caption{
    Distribution of images and questions in English by chart type in \textsc{PolyChartQA}.
  }
  \label{fig:engdisqa}
\end{figure*}

\begin{figure*}[htbp]
  \centering
  \includegraphics[width=\textwidth]{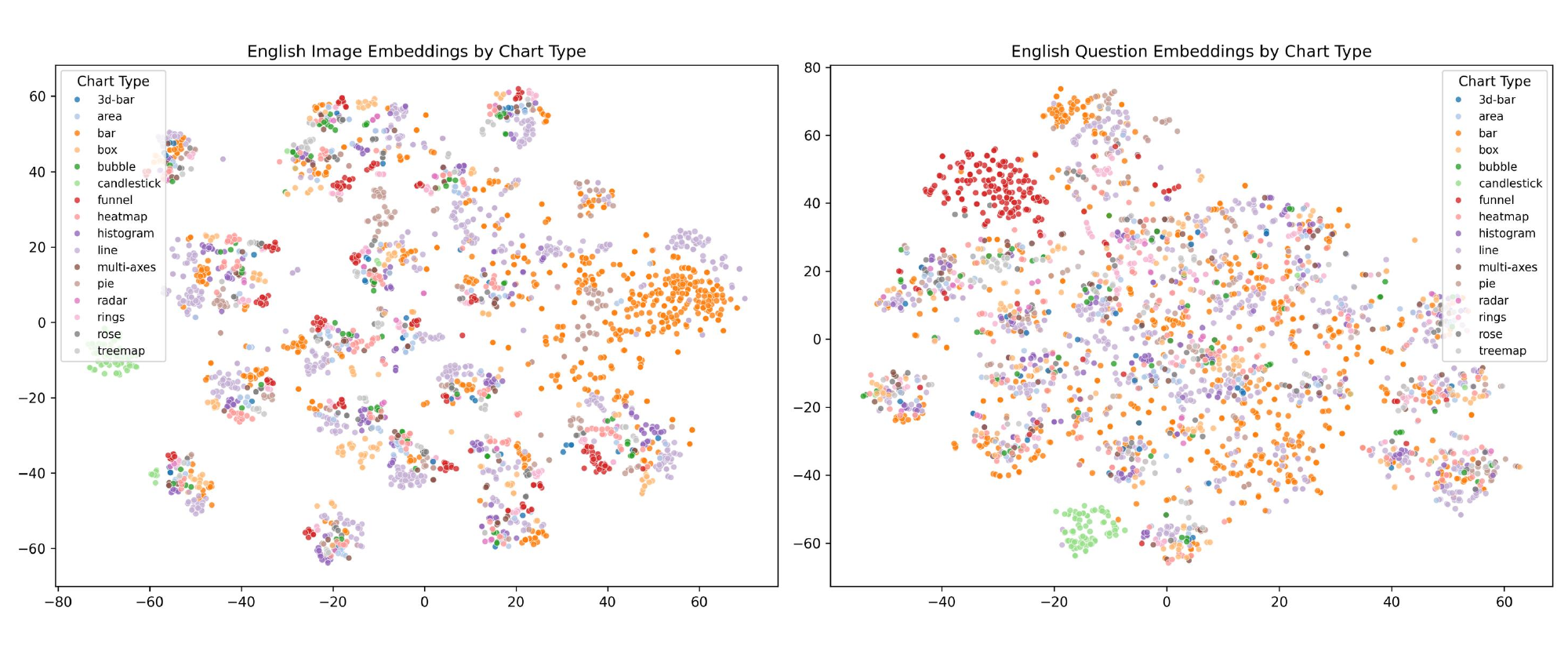}
  \caption{
    Distribution of JSON data and code in English by chart type in \textsc{PolyChartQA}.
  }
  \label{fig:engdisimg}
\end{figure*}

We also provide a detailed analysis of the English subset, which serves as the seed data for \textsc{PolyChartQA}. Figure~\ref{fig:engdisqa} shows t-SNE visualizations of image and question embeddings, with points colored by chart type to reveal clustering based on visual and semantic chart characteristics. Figure~\ref{fig:engdisimg} presents t-SNE plots of embeddings from the JSON data underlying the charts and the Python code used to generate them, again colored by chart type. These analyses illustrate the extent to which chart types can be distinguished within visual, textual, and structural representations.

\paragraph{Distribution of Images and Questions by Language.}

\begin{figure*}[htbp]
  \centering
  \includegraphics[width=\textwidth]{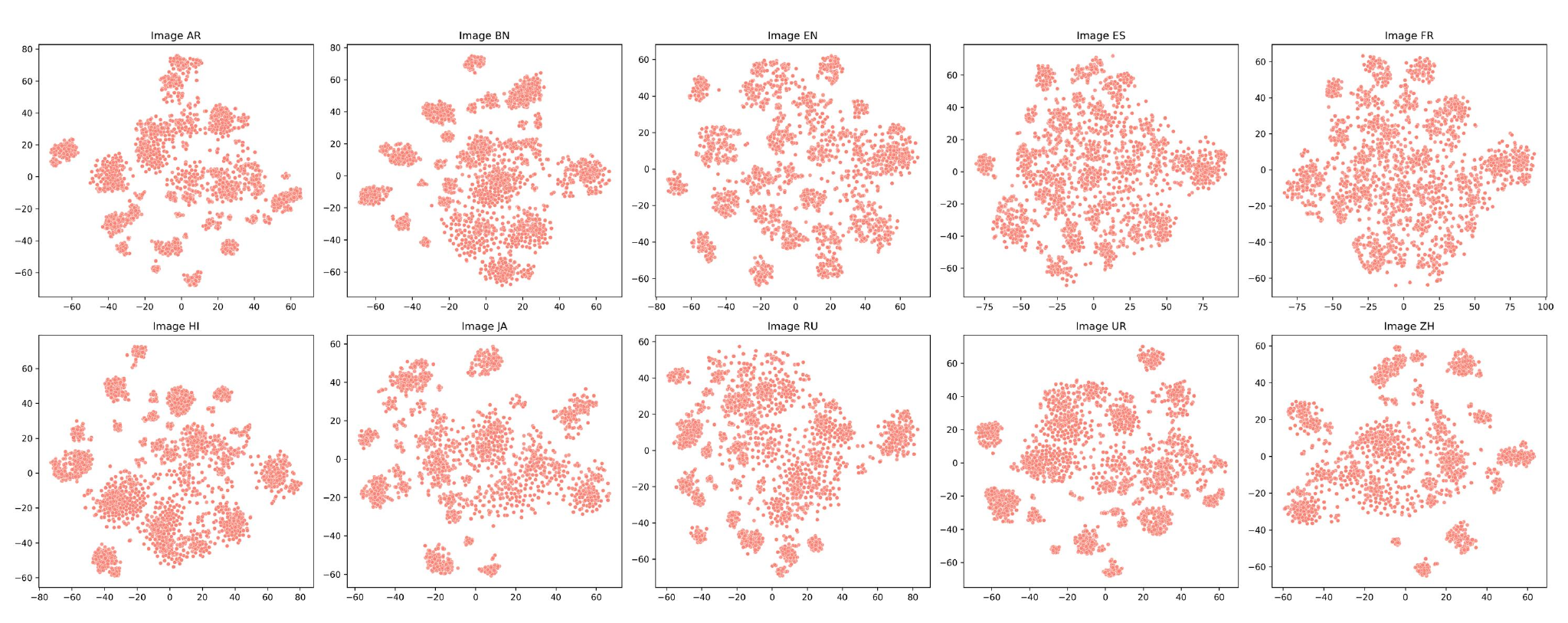}
  \caption{
    Distribution of images in \textsc{PolyChartQA} by language.
  }
  \label{fig:alldisimg}
\end{figure*}

\begin{figure*}[htbp]
  \centering
  \includegraphics[width=\textwidth]{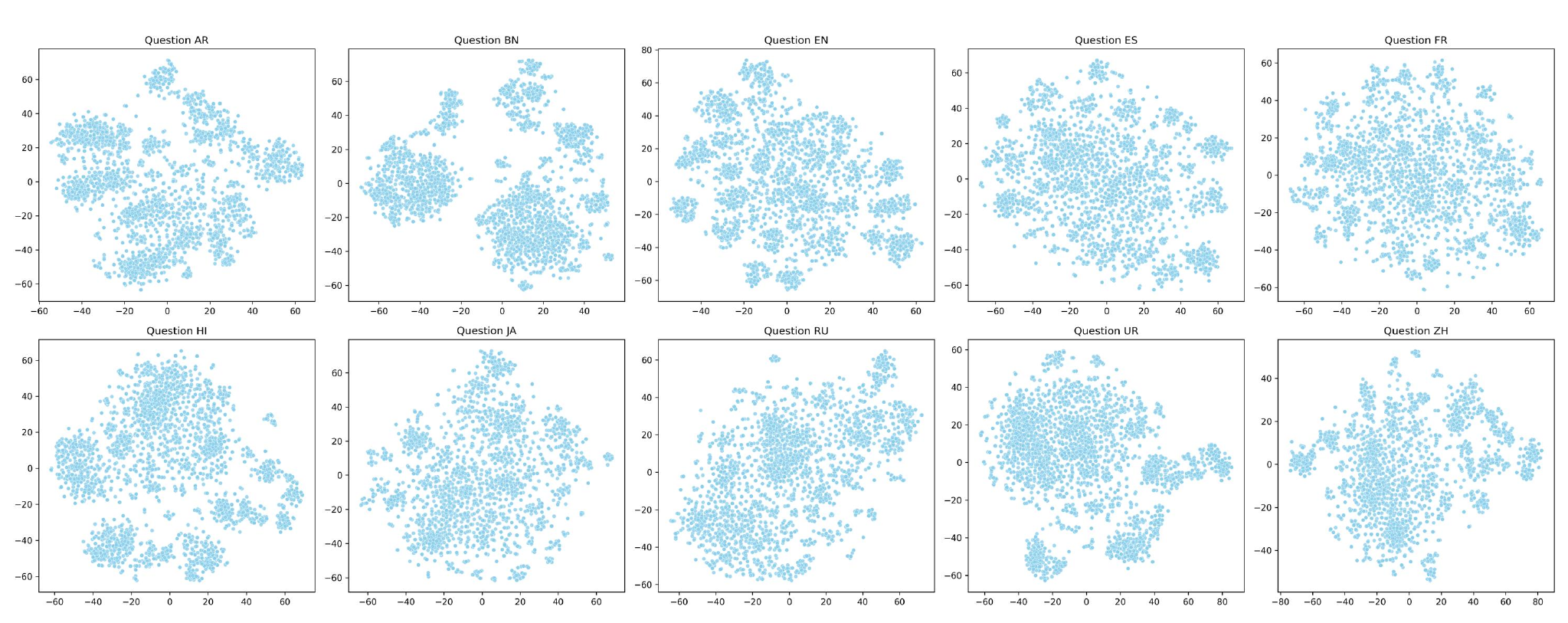}
  \caption{
    Distribution of questions in \textsc{PolyChartQA} by language.
  }
  \label{fig:alldisqa}
\end{figure*}

We further examine the distribution of images and questions in each language. Figure~\ref{fig:alldisimg} presents a t-SNE visualization of CLIP image embeddings, while Figure~\ref{fig:alldisqa} visualizes CLIP text embeddings of questions. In both cases, each subplot corresponds to a specific language. All points are uniformly colored to emphasize intra-language distribution rather than inter-category variation. These visualizations reveal the diversity and clustering patterns present in the multilingual data.

\subsection{\textsc{PolyChartQA}-Train}
\label{sec:appendix_polychartqa_train}

\paragraph{Data Statistics by Language and Chart Type}
We show the detailed statistics of \textsc{PolyChartQA}-Train in Tables~\ref{tab:polychartinstruct_image_stats} and~\ref{tab:polychartinstruct_qa_stats}, including per-language and per-chart-type breakdowns for both images and QA pairs.

\section{Human Evaluation Details}
\label{app:human_eval}

\subsection{Information of Human Annotators}

We conducted a rigorous human evaluation to measure the quality of multilingual chart images and their question-answering pairs in \textsc{PolyChartQA}. All annotators are either native speakers with over 15 years of experience in the target language or individuals holding a bachelor's degree and official certification in the corresponding language. We recruit two annotators for each language.

\subsection{Annotation Process}
All annotations were collected via crowdsourcing. Annotators reviewed HTML-rendered charts and questions, and recorded their responses in structured Excel spreadsheets. Full instructions provided to human annotators are detailed below. 
\begin{blackpromptbox}[title=Full Human Evaluation Instructions]
\textbf{Evaluation Dimensions \& Criteria:}

(1) Image Quality Assessment: Assess the visual quality of the target language chart. Evaluate its clarity, the legibility and correctness of all text and graphical elements, and its overall professional integrity.
\begin{itemize}[leftmargin=*]
    \item 3: The image is clear, professional, and undistorted. All text and graphical elements are correctly displayed and legible. The chart type accurately reflects the data.
    \item 2: The chart has minor flaws, such as slight blurriness or minor display issues, but these do not significantly hinder comprehension.
    \item 1: The chart has major issues (e.g., distortion, illegible text, incorrect chart type) that hinder or prevent comprehension.
\end{itemize}

(2) QA Correctness Assessment: Assess if the question is relevant to the chart and if the answer is factually correct and fully supported by the information presented in the target language chart.
\begin{itemize}[leftmargin=*]
    \item 3: The question is relevant, and the answer is correct and fully supported by the chart data.
    \item 2: The QA pair has minor errors or ambiguities. The question might be slightly unclear, or the answer may have small inaccuracies.
    \item 1: The question is irrelevant to the chart, or the answer is factually incorrect or unsupported by the chart.
\end{itemize}

(3) Translation Accuracy: Evaluate the quality of the image and QA translation from English to the target language. Assess its fidelity, semantic consistency, and natural fluency, and check if it conforms to the target language's idiomatic expressions. Crucially, determine if the translation introduces any bias, misinformation, or framing.
\begin{itemize}[leftmargin=*]
    \item 3: The translation is accurate, fluent, and natural, conforming perfectly to the target language's conventions. It preserves the original meaning and key information without introducing any bias, misinformation, or framing.
    \item 2: The translation is mostly correct and preserves the core meaning, but has minor issues like awkward phrasing or does not feel fully idiomatic. It may subtly introduce minor bias or framing, but does not significantly mislead.
    \item 1: The translation has major errors, is semantically inconsistent, or is highly unnatural. Additionally, or as a primary issue, it introduces clear bias, misinformation, or framing that distorts the original message.
\end{itemize}
\end{blackpromptbox}

\begin{figure*}[htbp]
  \centering
  \includegraphics[width=0.95\columnwidth]{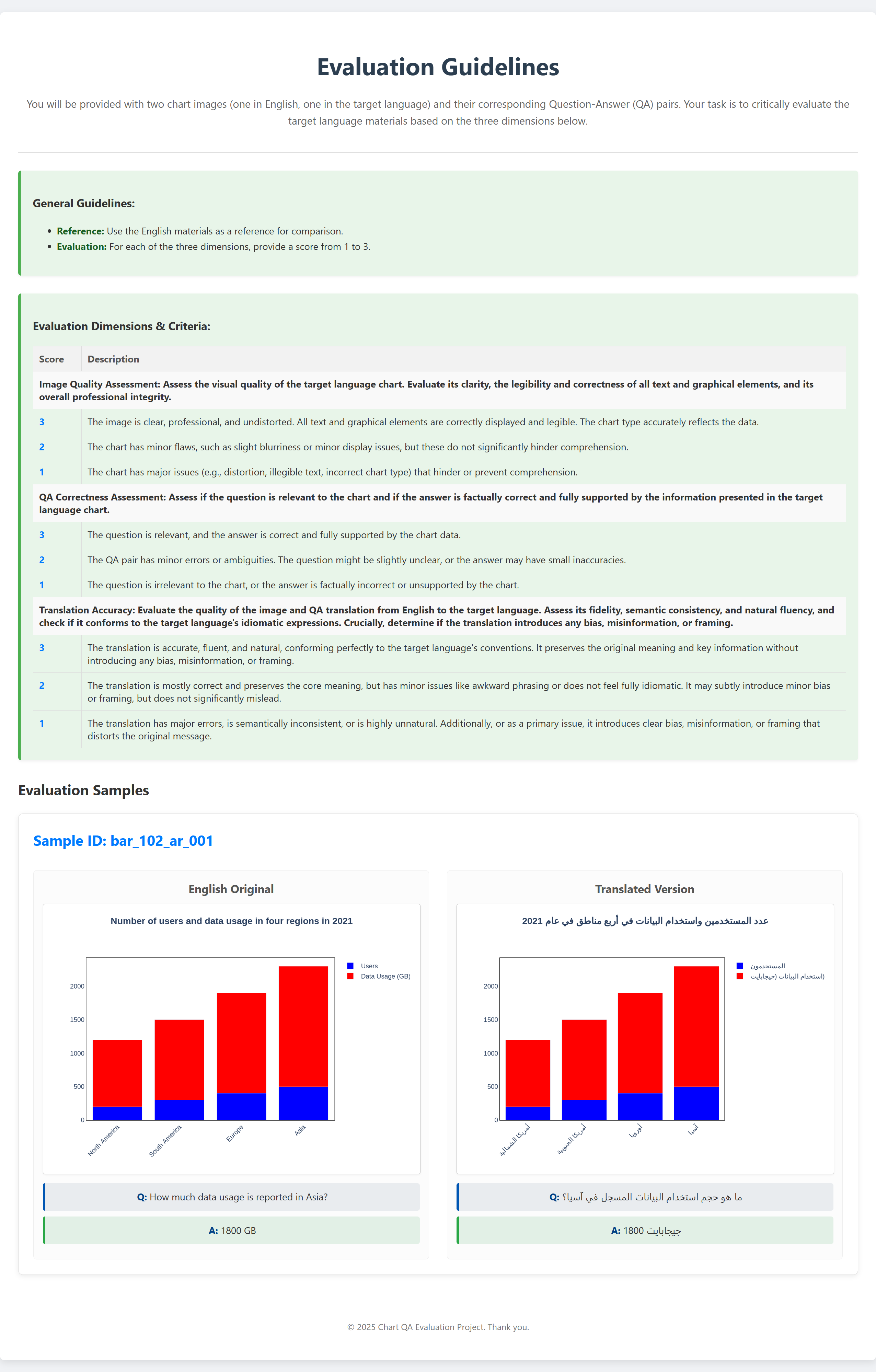}
  \caption{
    Human evaluation interface.
    Annotators review chart images and QA pairs in both source and target languages, providing quality ratings for image quality, QA correctness and translation accuracy.
  }
  \label{fig:human_eval_ui}
\end{figure*}

Figure~\ref{fig:human_eval_ui} shows an example of the custom annotation interface designed for this task, enabling annotators to efficiently compare original and translated chart images as well as their corresponding question-answer pairs.

\subsection{Annotation Results Details}
We present the complete results of human annotations in Table~\ref{tab:full_human_eval_appendix}. For each language, we report the average human score, inter-annotator agreement, and the weighted Cohen’s $\kappa$ between annotators. These consistently high scores indicate strong annotator consistency and confidence, further validating the overall quality and reliability of our dataset.
    
\begin{table*}[t!]
\centering
\begin{adjustbox}{max width=\textwidth}
\begin{tabular}{@{}l ccc ccc ccc@{}}
\toprule
\multirow{2}{*}{\textbf{Language}} & \multicolumn{3}{c}{\textbf{Image Quality}} & \multicolumn{3}{c}{\textbf{QA Relevance}} & \multicolumn{3}{c}{\textbf{Translation Accuracy}} \\
\cmidrule(lr){2-4} \cmidrule(lr){5-7} \cmidrule(lr){8-10}
& \textbf{Avg. Score} & \textbf{Disag.} & \textbf{$\kappa_w$} & \textbf{Avg. Score} & \textbf{Disag.} & \textbf{$\kappa_w$} & \textbf{Avg. Score} & \textbf{Disag.} & \textbf{$\kappa_w$} \\
\midrule
Arabic      & 2.94 & 2 & 0.929 & 2.97 & 5 & 0.656 & 2.79 & 3 & 0.964 \\
Urdu        & 2.71 & 3 & 0.971 & 2.92 & 4 & 0.891 & 2.71 & 7 & 0.932 \\
Hindi       & 2.93 & 3 & 0.908 & 3.00 & 0 & --\textsuperscript{*} & 2.95 & 3 & 0.874 \\
Bengali     & 2.91 & 7 & 0.829 & 2.98 & 2 & 0.796 & 2.92 & 6 & 0.837 \\
Chinese     & 2.96 & 2 & 0.896 & 2.98 & 2 & 0.796 & 2.95 & 3 & 0.874 \\
French      & 2.92 & 4 & 0.891 & 2.95 & 5 & 0.789 & 2.91 & 1 & 0.976 \\
Spanish     & 2.84 & 2 & 0.970 & 2.95 & 5 & 0.789 & 2.87 & 5 & 0.912 \\
Russian     & 2.65 & 5 & 0.956 & 2.71 & 3 & 0.971 & 2.92 & 2 & 0.946 \\
Japanese    & 2.86 & 2 & 0.967 & 2.90 & 6 & 0.867 & 2.95 & 5 & 0.789 \\
English     & 2.95 & 1 & 0.958 & 2.98 & 2 & 0.796 & 2.96 & 4 & 0.792 \\
\midrule
\textbf{Average} & \textbf{2.87} & \textbf{3.6} & \textbf{0.927} & \textbf{2.93} & \textbf{3.9} & \textbf{0.817} & \textbf{2.89} & \textbf{3.9} & \textbf{0.889} \\
\bottomrule
\multicolumn{10}{l}{\textsuperscript{*}\footnotesize{Kappa is undefined due to zero variance (100\% agreement). This entry was excluded from the average calculation.}}
\end{tabular}
\end{adjustbox}
\caption{Detailed human scores and inter-annotator agreement scores for each language and evaluation dimension. Scores are based on 250 items per language rated by two annotators. "Disag." shows the raw count of differing ratings and $\kappa_w$ denotes weighted Cohen’s $\kappa$.}
\label{tab:full_human_eval_appendix}
\end{table*}

\section{More Implementation Details}
\label{app:eval_detail}

\subsection{Metric Details}
For METEOR metric, we use its official code from huggingface\footnote{\url{https://huggingface.co/spaces/evaluate-metric/meteor}}.

\subsection{Models Details}
The general open-source LVLMs include Qwen2-VL~\cite{wang2024qwen2}, Qwen2.5-VL~\cite{bai2025qwen2}, InternVL-2.5~\cite{chen2024expanding}, InternVL-3~\cite{zhu2025internvl3}, Phi-3 Vision~\cite{abdin2024phi}, Phi-4 Multimodal~\cite{abdin2024phi_4}, PaliGemma 2~\cite{team2024gemma_2}, LLaVA-v1.6~\cite{llava1_6}, LLaVA-OneVision~\cite{llava_one_vision}, Llama-3.2-Vision~\cite{llama_3_2_vision}, and DeepSeek-VL2~\cite{wu2024deepseekvl2mixtureofexpertsvisionlanguagemodels}. For open-source multilingual LVLMs, we evaluate PALO~\cite{maaz2024palo}, Maya~\cite{alam2024mayainstructionfinetunedmultilingual}, Pangea~\cite{yue2024pangea}, and Centurio~\cite{geigle2025centurio}. The chart-specific category includes TinyChart~\cite{zhang2024tinychart}, ChartGemma~\cite{masry2025chartgemma}, ChartInstruct~\cite{masry2024chartinstruct}, ChartLlama~\citep{han2023chartllama}, and ChartAssistant~\citep{meng2024chartassisstant}. Closed-source category comprises Gemini-2.5-Pro~\cite{comanici2025gemini} and GPT-4o~\cite{gpt4o}. Closed-source models are accessed via their official APIs, while open-source models are run using their instruct versions available on the Hugging Face Model Hub.

\section{More Experiments}
\label{app:experiments}


We further conduct a series of experiments on model inference and training, including ablations on English data ratio (§\ref{app:more_exp_abl_english_ratio}), two-stage post-training (§\ref{app:more_exp_two_stage}) and fine-tuning settings. These analyses reveal key insights into the weaknesses of current models and provide guidance for improving their multilingual chart understanding capabilities. We also present the complete experimental results corresponding to the main paper, including fine-tuning results (§\ref{app:more_exp_ft}), and ablation on training data percentage (§\ref{app:more_exp_abl_data_percentage}).




\subsection{Ablation on English Data Ratio}
\label{app:more_exp_abl_english_ratio}

\begin{figure}[htbp]

  \includegraphics[width=\columnwidth]{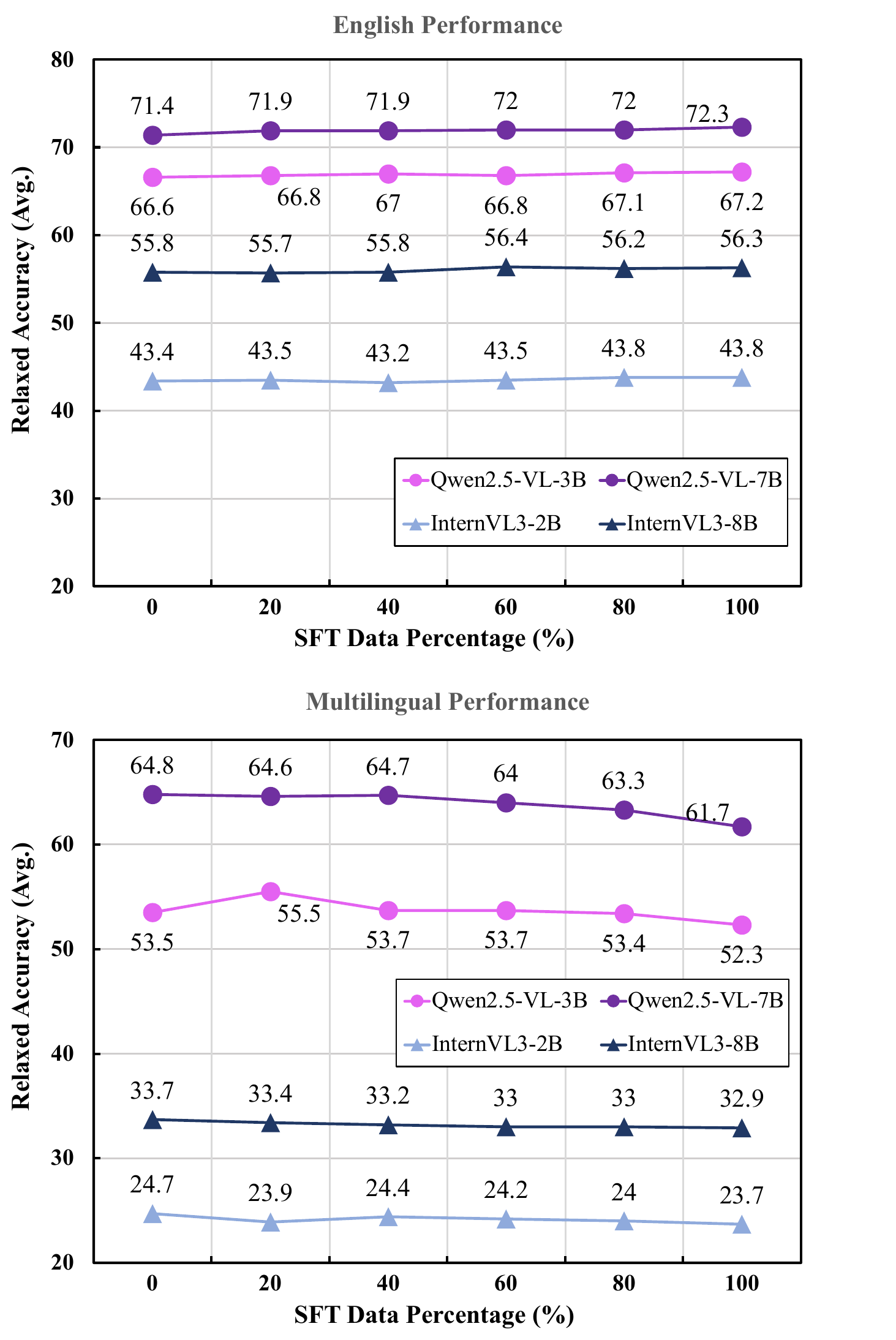}
  \caption{
   Performance on \textsc{PolyChartQA} with respect to the English data ratio across different model families.
  }
  \vspace{-5pt}
  \label{fig:ablation_eng_ratio}
\end{figure}

To investigate the impact of English data proportion in multilingual fine-tuning, we conduct an ablation study by varying the ratio of English samples from 0\% to 100\% while keeping the total dataset size fixed at 70K QA pairs. The remaining proportion (i.e., non-English data) is evenly distributed across the other nine languages to ensure balanced multilingual representation. As shown in Figure~\ref{fig:ablation_eng_ratio} and Table~\ref{tab:model_accuracy_english_ratio}, increasing the proportion of English data does not consistently enhance multilingual performance. Larger models like Qwen2.5-VL-7B maintain stable accuracy across all ratios, suggesting strong multilingual robustness, whereas smaller models such as InternVL3 exhibit slight degradation when English data dominates, likely due to reduced exposure to multilingual contexts. Overall, excessive reliance on English offers limited benefit and may even weaken cross-lingual generalization.

\subsection{Two-stage Fine-tuning on Qwen-2.5-VL}
\label{app:more_exp_two_stage}

In this section, we investigate whether the multilingual chart understanding ability of models can be further improved through a two-stage training strategy. We choose Qwen2.5-VL as our base model. In the first stage, we construct an alignment dataset using \textsc{PolyChartQA}-Train and other open-source resources. We then perform alignment training followed by fine-tuning on \textsc{PolyChartQA}-Train. Additionally, we examine the impact of unfreezing the vision encoder in each stage on overall performance. We further discuss the results and provide training insights below.

\paragraph{Data Construction for Alignment Stage}

In the alignment stage, we aim to achieve multilingual alignment through a chart-to-JSON prediction task using the chart metadata from \textsc{PolyChartQA}-Train. To further strengthen multilingual visual–textual grounding, we incorporate additional document and chart OCR tasks from external datasets, including MTVQA, PangeaOCR, and SMPQA. In total, this stage involves approximately 850K samples, comprising:

\begin{enumerate}
    \item \textsc{PolyChartQA}-Train. We extract image–JSON pairs from \textsc{PolyChartQA}-Train, yielding approximately 131K instances.
    \item \textbf{MTVQA.} We incorporate the full training split of MTVQA~\cite{tang2024mtvqa}, which contains 21K chart–QA pairs.
    \item \textbf{Pangea.} We include 300K OCR data samples from the Pangea-OCR dataset~\cite{yue2024pangea}.
    \item \textbf{SMPQA-Reconstructed.} Following \citet{geigle2025centurio}, we adapt SMPQA to our 10-language setting by reconstructing 410K synthetic chart-OCR training examples.
\end{enumerate}

\paragraph{Two-stage Training Results}

We apply LoRA~\cite{hu2022lora} in both stages with a fixed $r=128$. The alignment stage uses a learning rate of $5e^-5$, while the instruction tuning stage uses a learning rate of $1e^-5$. Each stage is trained for one epoch.

Table~\ref{tab:ve_training_strategies_3b} and Table~\ref{tab:ve_training_strategies_full_x100} present the full ablation results of Qwen2.5-VL-3B and 7B, respectively. Across both model sizes, we observe consistent patterns: (i) fine-tuning alone provides substantial gains over the baseline, and (ii) incorporating an additional alignment stage further improves performance. Notably, the configuration where the vision encoder is unfrozen during alignment but frozen during instruction tuning achieves the highest accuracy in both models (63.6 for 3B, 68.0 for 7B). These results confirm that gradual visual adaptation followed by stabilization is a robust strategy for enhancing multilingual chart understanding across different model scales. This also indicates that the ability of models to understand multilingual charts can be further enhanced through additional training strategies.

\begin{table*}[htbp]
\centering
\begin{adjustbox}{width=\textwidth,center}
\begin{tabular}{lcccccccccc|c}
\toprule
\textbf{Chart Type} & \textbf{EN} & \textbf{AR} & \textbf{BN} & \textbf{ES} & \textbf{FR} & \textbf{HI} & \textbf{JA} & \textbf{RU} & \textbf{UR} & \textbf{ZH} & \textbf{Total} \\
\midrule
3d-bar              & 40 & 31 & 27 & 35 & 35 & 30 & 26 & 30 & 30 & 26 & \textbf{310} \\
area                & 106 & 79 & 76 & 84 & 78 & 86 & 61 & 65 & 68 & 63 & \textbf{766} \\
bar                 & 600 & 447 & 507 & 505 & 471 & 547 & 409 & 477 & 514 & 393 & \textbf{4870} \\
box                 & 171 & 144 & 155 & 148 & 144 & 153 & 131 & 132 & 153 & 134 & \textbf{1465} \\
bubble              & 81 & 32 & 39 & 38 & 38 & 40 & 33 & 35 & 37 & 35 & \textbf{408} \\
candlestick         & 86 & 62 & 67 & 74 & 62 & 70 & 50 & 56 & 61 & 56 & \textbf{644} \\
funnel              & 211 & 148 & 155 & 158 & 154 & 165 & 121 & 142 & 137 & 117 & \textbf{1508} \\
heatmap             & 183 & 133 & 149 & 149 & 153 & 160 & 120 & 134 & 153 & 125 & \textbf{1459} \\
histogram           & 219 & 167 & 177 & 180 & 187 & 182 & 141 & 162 & 181 & 137 & \textbf{1733} \\
line                & 600 & 491 & 500 & 551 & 521 & 539 & 436 & 516 & 509 & 402 & \textbf{5065} \\
multi-axes          & 77 & 49 & 53 & 52 & 58 & 55 & 42 & 48 & 58 & 45 & \textbf{537} \\
pie                 & 190 & 133 & 148 & 150 & 148 & 162 & 120 & 130 & 146 & 93 & \textbf{1420} \\
radar               & 42 & 23 & 25 & 26 & 24 & 29 & 27 & 24 & 26 & 21 & \textbf{267} \\
rings               & 123 & 80 & 83 & 91 & 95 & 92 & 72 & 66 & 85 & 76 & \textbf{863} \\
rose                & 84 & 46 & 58 & 53 & 61 & 64 & 36 & 44 & 54 & 34 & \textbf{534} \\
treemap             & 104 & 74 & 78 & 85 & 75 & 78 & 68 & 63 & 72 & 60 & \textbf{757} \\
\midrule
\textbf{Total}      & \textbf{2917} & \textbf{2139} & \textbf{2297} & \textbf{2379} & \textbf{2304} & \textbf{2452} & \textbf{1893} & \textbf{2124} & \textbf{2284} & \textbf{1817} & \textbf{22606} \\
\bottomrule
\end{tabular}
\end{adjustbox}
\caption{Detailed statistics of Image counts per chart type across all languages in \textsc{PolyChartQA}. 
}
\label{tab:polychartqa_image_stats}
\end{table*}

\begin{table*}[htbp]
\centering
\begin{adjustbox}{width=\textwidth,center}
\begin{tabular}{lcccccccccc|c}
\toprule
\textbf{Chart Type} & \textbf{EN} & \textbf{AR} & \textbf{BN} & \textbf{ES} & \textbf{FR} & \textbf{HI} & \textbf{JA} & \textbf{RU} & \textbf{UR} & \textbf{ZH} & \textbf{Total} \\
\midrule
3d-bar              & 40  & 31  & 27  & 35  & 35  & 30  & 26  & 30  & 30  & 26  & \textbf{310} \\
area                & 107 & 80  & 77  & 85  & 79  & 87  & 62  & 66  & 69  & 64  & \textbf{776} \\
bar                 & 696 & 592 & 670 & 669 & 627 & 733 & 535 & 638 & 685 & 517 & \textbf{6362} \\
box                 & 171 & 144 & 155 & 148 & 144 & 153 & 131 & 132 & 153 & 134 & \textbf{1465} \\
bubble              & 81  & 32  & 39  & 38  & 38  & 40  & 33  & 35  & 37  & 35  & \textbf{408} \\
candlestick         & 86  & 62  & 67  & 74  & 62  & 70  & 50  & 56  & 61  & 56  & \textbf{644} \\
funnel              & 211 & 148 & 155 & 158 & 154 & 165 & 121 & 142 & 137 & 117 & \textbf{1508} \\
heatmap             & 183 & 133 & 149 & 149 & 153 & 160 & 120 & 134 & 153 & 125 & \textbf{1459} \\
histogram           & 219 & 167 & 177 & 180 & 187 & 182 & 141 & 162 & 181 & 137 & \textbf{1733} \\
line                & 646 & 689 & 718 & 794 & 739 & 770 & 602 & 734 & 720 & 551 & \textbf{6963} \\
multi-axes          & 77  & 49  & 53  & 52  & 58  & 55  & 42  & 48  & 58  & 45  & \textbf{537} \\
pie                 & 210 & 146 & 164 & 165 & 163 & 178 & 129 & 145 & 159 & 106 & \textbf{1565} \\
radar               & 42  & 23  & 25  & 26  & 24  & 29  & 27  & 24  & 26  & 21  & \textbf{267} \\
rings               & 123 & 80  & 83  & 91  & 95  & 92  & 72  & 66  & 85  & 76  & \textbf{863} \\
rose                & 84  & 46  & 58  & 53  & 61  & 64  & 36  & 44  & 54  & 34  & \textbf{534} \\
treemap             & 104 & 74  & 78  & 85  & 75  & 78  & 68  & 63  & 72  & 60  & \textbf{757} \\
\midrule
\textbf{Total}      & \textbf{3080} & \textbf{2496} & \textbf{2695} & \textbf{2802} & \textbf{2694} & \textbf{2886} & \textbf{2195} & \textbf{2519} & \textbf{2680} & \textbf{2104} & \textbf{26151} \\
\bottomrule
\end{tabular}
\end{adjustbox}
\caption{Detailed statistics of Question-Answer (QA) pair counts per chart type across all languages in \textsc{PolyChartQA}
}
\label{tab:polychartqa_qa_stats}
\end{table*}

\begin{table*}[htbp]
\centering
\begin{adjustbox}{width=\textwidth,center}
\begin{tabular}{lcccccccccc|c}
\toprule
\textbf{Chart Type} & \textbf{AR} & \textbf{BN} & \textbf{EN} & \textbf{ES} & \textbf{FR} & \textbf{HI} & \textbf{JA} & \textbf{RU} & \textbf{UR} & \textbf{ZH} & \textbf{Total} \\
\midrule
3d-bar              & 4    & 4    & 4    & 2    & 3    & 3    & 3    & 2    & 4    & 4    & \textbf{33} \\
area                & 1    & 1    & 1    & 1    & 1    & 1    & 1    & 1    & 1    & 1    & \textbf{10} \\
bar                 & 7834 & 7978 & 8876 & 7726 & 7878 & 8049 & 7883 & 7955 & 7804 & 8000 & \textbf{79983} \\
box                 & 50   & 47   & 57   & 47   & 48   & 46   & 46   & 48   & 49   & 50   & \textbf{488} \\
candlestick         & 231  & 224  & 267  & 226  & 240  & 244  & 222  & 223  & 223  & 231  & \textbf{2331} \\
funnel              & 103  & 107  & 118  & 101  & 96   & 107  & 102  & 106  & 100  & 102  & \textbf{1042} \\
gantt               & 110  & 101  & 143  & 122  & 119  & 122  & 114  & 117  & 99   & 114  & \textbf{1161} \\
heatmap             & 154  & 160  & 218  & 162  & 155  & 167  & 168  & 169  & 174  & 153  & \textbf{1680} \\
line                & 3281 & 3383 & 3937 & 3220 & 3281 & 3374 & 3294 & 3374 & 3348 & 3340 & \textbf{33832} \\
other               & 13   & 17   & 17   & 14   & 16   & 17   & 16   & 15   & 14   & 13   & \textbf{152} \\
pie                 & 630  & 629  & 781  & 602  & 593  & 645  & 632  & 643  & 631  & 630  & \textbf{6416} \\
radar               & 184  & 176  & 203  & 166  & 165  & 185  & 165  & 177  & 167  & 173  & \textbf{1761} \\
rings               & 66   & 68   & 88   & 67   & 69   & 68   & 68   & 69   & 68   & 70   & \textbf{699} \\
scatter             & 186  & 193  & 222  & 182  & 185  & 185  & 188  & 200  & 187  & 196  & \textbf{1924} \\
\midrule
\textbf{Total}      & \textbf{12847} & \textbf{13088} & \textbf{14932} & \textbf{12638} & \textbf{12849} & \textbf{13213} & \textbf{12902} & \textbf{13099} & \textbf{12869} & \textbf{13078} & \textbf{131515} \\
\bottomrule
\end{tabular}
\end{adjustbox}
\caption{Detailed statistics of Image counts per chart type across all languages in \textsc{PolyChartQA}-Train. 
}
\label{tab:polychartinstruct_image_stats}
\end{table*}

\begin{table*}[htbp]
\centering
\begin{adjustbox}{width=\textwidth,center}
\begin{tabular}{lcccccccccc|c}
\toprule
\textbf{Chart Type} & \textbf{AR} & \textbf{BN} & \textbf{EN} & \textbf{ES} & \textbf{FR} & \textbf{HI} & \textbf{JA} & \textbf{RU} & \textbf{UR} & \textbf{ZH} & \textbf{Total} \\
\midrule
3d-bar              & 41    & 41    & 41    & 21    & 30    & 31    & 30    & 20    & 41    & 41    & \textbf{317} \\
area                & 1     & 1     & 1     & 1     & 1     & 1     & 1     & 1     & 1     & 1     & \textbf{10} \\
bar                 & 33161 & 33764 & 38339 & 32794 & 33463 & 33940 & 33385 & 33626 & 32962 & 33998 & \textbf{339432} \\
box                 & 510   & 479   & 580   & 478   & 491   & 467   & 468   & 488   & 500   & 510   & \textbf{4971} \\
candlestick         & 2279  & 2209  & 2639  & 2231  & 2369  & 2409  & 2190  & 2208  & 2201  & 2288  & \textbf{23023} \\
funnel              & 1055  & 1097  & 1202  & 1042  & 982   & 1096  & 1055  & 1086  & 1009  & 1044  & \textbf{10724} \\
gantt               & 1098  & 1008  & 1428  & 1218  & 1188  & 1219  & 1139  & 1169  & 989   & 1138  & \textbf{11592} \\
heatmap             & 1547  & 1609  & 2190  & 1629  & 1558  & 1679  & 1688  & 1698  & 1750  & 1539  & \textbf{16887} \\
line                & 25110 & 25942 & 30793 & 24645 & 25110 & 26013 & 25196 & 25998 & 25539 & 25763 & \textbf{260109} \\
other               & 98    & 129   & 129   & 91    & 128   & 129   & 119   & 109   & 98    & 115   & \textbf{1145} \\
pie                 & 3833  & 3779  & 4901  & 3617  & 3615  & 3947  & 3863  & 3909  & 3806  & 3856  & \textbf{39126} \\
radar               & 1845  & 1766  & 2042  & 1660  & 1662  & 1860  & 1663  & 1774  & 1671  & 1745  & \textbf{17688} \\
rings               & 669   & 688   & 893   & 680   & 700   & 688   & 691   & 700   & 688   & 711   & \textbf{7108} \\
scatter             & 1857  & 1933  & 2221  & 1824  & 1856  & 1857  & 1885  & 1997  & 1880  & 1955  & \textbf{19265} \\
\midrule
\textbf{Total}      & \textbf{73104} & \textbf{74445} & \textbf{87399} & \textbf{71931} & \textbf{73153} & \textbf{75336} & \textbf{73373} & \textbf{74783} & \textbf{73135} & \textbf{74704} & \textbf{751363} \\
\bottomrule
\end{tabular}
\end{adjustbox}
\caption{Detailed statistics of QA pair counts per chart type across all languages in \textsc{PolyChartQA}-Train.
}
\label{tab:polychartinstruct_qa_stats}
\end{table*}

\begin{table*}[t]
\small
\centering
\setlength{\tabcolsep}{4pt}
\begin{adjustbox}{max width=\textwidth}
\begin{tabular}{@{}l|c|cccccccccc|cc@{}}
\toprule
\textbf{Model} & \textbf{\% EN Data} & {\textbf{EN}} & {\textbf{ZH}} & {\textbf{FR}} & {\textbf{ES}} & {\textbf{RU}} & {\textbf{JA}} & {\textbf{AR}} & {\textbf{HI}} & {\textbf{UR}} & {\textbf{BN}} & {\textbf{Avg. (w EN)}} & {\textbf{Avg. (w/o EN)}} \\
\midrule
\multirow{6}{*}{\textbf{InternVL3-2B}} 
& 0 & 43.4 & \textbf{35.9} & \textbf{34.3} & \textbf{36.0} & 29.2 & \textbf{26.3} & \textbf{18.3} & \underline{16.5} & \textbf{15.6} & \textbf{13.1} & \textbf{26.9} & \textbf{24.7} \\
& 20 & \underline{43.5} & 33.1 & 32.1 & 35.6 & \underline{29.6} & 24.4 & \textbf{18.3} & 16.2 & 15.4 & 12.4 & 26.2 & 23.9 \\
& 40 & 43.2 & 34.8 & \underline{33.9} & \underline{35.8} & 29.4 & \underline{25.8} & 18.2 & \textbf{16.3} & \underline{15.7} & 12.6 & \underline{26.6} & \underline{24.4} \\
& 60 & 43.5 & 34.5 & 33.5 & 35.5 & 29.1 & 25.0 & 18.1 & 16.0 & 15.4 & \underline{12.8} & 26.4 & 24.2 \\
& 80 & \textbf{43.8} & 33.7 & 33.1 & 35.8 & 28.8 & 24.1 & \underline{18.2} & 16.0 & 15.6 & 12.7 & 26.3 & 24.0 \\
& 100 & \textbf{43.8} & \underline{32.9} & 32.3 & 35.3 & 28.9 & 23.8 & \underline{18.2} & 15.9 & 15.6 & 12.5 & 26.1 & 23.7 \\
\midrule
\multirow{6}{*}{\textbf{InternVL3-8B}} 
& 0 & 55.8 & \textbf{45.4} & \textbf{47.6} & \textbf{51.0} & \textbf{41.3} & \textbf{40.3} & \textbf{22.4} & \textbf{21.3} & \textbf{18.4} & \underline{19.1} & \textbf{36.3} & \textbf{33.7} \\
& 20 & 55.7 & 44.8 & 47.0 & 50.5 & \textbf{41.4} & 40.0 & 22.2 & 21.2 & 18.2 & \textbf{19.1} & 36.1 & 33.4 \\
& 40 & 55.8 & 43.8 & 46.4 & 50.4 & 41.3 & 39.7 & 22.3 & 21.1 & 18.1 & 18.7 & 35.8 & 33.2 \\
& 60 & \underline{56.4} & 42.9 & 46.3 & 50.3 & 41.4 & 39.5 & 22.1 & 21.1 & \textbf{18.2} & 18.6 & \underline{35.8} & 33.0 \\
& 80 & 56.2 & 42.0 & 46.1 & 50.2 & 41.4 & 39.7 & 22.1 & 21.3 & \underline{18.2} & 18.6 & 35.7 & 33.0 \\
& 100 & \textbf{56.3} & \underline{40.8} & \underline{46.4} & \underline{50.2} & 41.0 & \underline{40.0} & \underline{22.3} & \underline{21.3} & 18.0 & \underline{18.7} & 35.6 & \underline{32.9} \\
\midrule
\multirow{6}{*}{\textbf{Qwen2.5-VL-3B}} 
& 0 & 66.6 & 60.6 & 63.0 & 62.7 & 59.7 & 53.8 & \textbf{54.0} & 47.2 & \textbf{39.6} & 43.1 & 55.0 & 53.5 \\
& 20 & 66.8 & \textbf{61.7} & \textbf{63.9} & 62.8 & \textbf{61.9} & \textbf{57.3} & \textbf{55.5} & \textbf{50.5} & \underline{42.7} & \textbf{45.4} & \textbf{56.8} & \textbf{55.5} \\
& 40 & \underline{67.0} & 60.5 & 63.5 & 62.6 & 60.6 & 53.6 & 54.0 & 47.6 & 40.2 & 43.0 & 55.3 & 53.7 \\
& 60 & 66.8 & 60.8 & 63.4 & 62.8 & 61.0 & 53.0 & 53.9 & 47.6 & 40.0 & 43.0 & 55.3 & 53.7 \\
& 80 & 67.1 & 60.7 & 63.7 & \underline{63.2} & 61.1 & 51.1 & 53.4 & \underline{47.4} & 38.7 & 42.6 & \underline{55.0} & \underline{53.4} \\
& 100 & \textbf{67.2} & \underline{59.6} & \underline{63.4} & 63.2 & \underline{60.3} & \underline{48.6} & \underline{51.6} & 46.4 & 37.1 & \underline{42.0} & 54.1 & 52.3 \\
\midrule
\multirow{6}{*}{\textbf{Qwen2.5-VL-7B}} 
& 0 & 71.4 & \textbf{68.0} & 70.2 & \textbf{69.5} & \textbf{68.9} & \textbf{66.2} & \textbf{63.4} & \textbf{62.4} & \textbf{56.6} & \textbf{58.8} & \textbf{65.5} & \textbf{64.8} \\
& 20 & \underline{71.9} & \underline{68.3} & 69.5 & 69.4 & 67.9 & \underline{67.2} & 63.1 & 62.1 & 56.2 & 58.8 & 65.4 & 64.6 \\
& 40 & 71.9 & 67.9 & \textbf{70.5} & \underline{69.7} & 68.1 & 66.5 & \underline{63.1} & \underline{62.3} & 56.3 & 58.6 & \underline{65.5} & \underline{64.7} \\
& 60 & 72.0 & 67.3 & 69.9 & 69.8 & 67.6 & 66.1 & 62.2 & 61.0 & 54.9 & 58.5 & 64.9 & 64.0 \\
& 80 & 72.0 & 66.5 & 69.2 & 69.4 & 67.2 & 64.6 & 61.2 & 60.6 & 54.3 & 57.5 & 64.3 & 63.3 \\
& 100 & \textbf{72.3} & 63.5 & \underline{69.7} & 69.4 & 67.1 & 59.6 & 61.5 & 58.6 & 51.1 & 54.8 & 62.9 & 61.7 \\
\bottomrule
\end{tabular}
\end{adjustbox}
\caption{Overall performance on the \textsc{PolyChartQA} benchmark under different \textbf{English data ratios}. For each model category, the best score per column is in \textbf{bold} and the second-best is \underline{underlined}.}
\label{tab:model_accuracy_english_ratio}
\end{table*}

\begin{table*}[htbp]
\centering
\resizebox{\textwidth}{!}{
\setlength{\tabcolsep}{3pt} 

\begin{tabular*}{\textwidth}{@{\extracolsep{\fill}} c cc *{10}{S[table-format=2.1]} S[table-format=2.1] S[table-format=2.1] @{}}
\toprule
\textbf{\begin{tabular}{@{}c@{}}Training\\Strategy\end{tabular}} & \textbf{Stage1} & \textbf{Stage2} & \textbf{EN} & \textbf{ZH} & \textbf{FR} & \textbf{ES} & \textbf{RU} & \textbf{JA} & \textbf{AR} & \textbf{UR} & \textbf{HI} & \textbf{BN} & \textbf{\begin{tabular}{@{}c@{}}Avg.\\(w EN)\end{tabular}} & \textbf{\begin{tabular}{@{}c@{}}Avg.\\(w/o EN)\end{tabular}} \\
\midrule
\multirow{1}{*}{\textit{Baseline}}
& \ding{55} & \ding{55}           & 67.4 & 59.6 & 61.8 & 62.5 & 58.0 & 48.8 & 51.4 & 37.2 & 45.7 & 43.0 & 53.7 & 51.8 \\
\cmidrule{2-15} 
\multirow{2}{*}{\textit{SFT only}}
& \ding{55} & {\small \faSnowflake} & 68.2 & 64.1 & 66.1 & 65.4 & 64.9 & 63.1 & 59.0 & 49.8 & 56.8 & 54.0 & 61.1 & 60.2 \\
& \ding{55} & {\small \faFire}      & 68.2 & 64.0 & \textbf{66.3} & 65.9 & 65.0 & 63.3 & 60.7 & 51.5 & 58.8 & 55.5 & 61.9 & 61.1 \\
\cmidrule{2-15} 
\multirow{3}{*}{\textit{Align+SFT}}
& {\small \faSnowflake}   & {\small \faSnowflake}   & 68.8 & 64.2 & 66.1 & \textbf{66.2} & 64.3 & 62.7 & 61.3 & 53.4 & 57.9 & 53.5 & 61.9 & 60.9 \\
& {\small \faFire} &  {\small \faSnowflake}  & 69.0 & \textbf{64.8} & 65.5 & \textbf{66.2} & 65.2 & \textbf{64.5} & \textbf{63.9} & \textbf{56.5} & \textbf{61.3} & \textbf{58.4} & \textbf{63.6} & \textbf{62.8} \\
& {\small \faFire} & {\small \faFire} & \textbf{69.3} & 64.1 & 64.9 & 66.0 & \textbf{65.5} & \textbf{64.5} & 63.8 & 55.6 & 61.1 & \textbf{58.4} & 63.4 & 62.6 \\
\bottomrule
\end{tabular*}
}
\caption{
Performance of different training strategies on Qwen2.5-VL-3B across various languages. {\small \faSnowflake} and {\small \faFire} indicate that the vision encoder is frozen or unfrozen, respectively, during each stage. \ding{55} denotes that the stage is skipped. Bold values denote the best performance.
}
\label{tab:ve_training_strategies_3b}
\end{table*}

\begin{table*}[htbp] 
\centering
\resizebox{\textwidth}{!}{
\setlength{\tabcolsep}{3pt} 

\begin{tabular*}{\textwidth}{@{\extracolsep{\fill}} c cc *{10}{S[table-format=2.1]} S[table-format=2.1] S[table-format=2.1] @{}}
\toprule
\textbf{\begin{tabular}{@{}c@{}}Training\\Strategy\end{tabular}} & \textbf{Stage1} & \textbf{Stage2} & \textbf{EN} & \textbf{ZH} & \textbf{FR} & \textbf{ES} & \textbf{RU} & \textbf{JA} & \textbf{AR} & \textbf{UR} & \textbf{HI} & \textbf{BN} & \textbf{\begin{tabular}{@{}c@{}}Avg.\\(w EN)\end{tabular}} & \textbf{\begin{tabular}{@{}c@{}}Avg.\\(w/o EN)\end{tabular}} \\
\midrule
\multirow{1}{*}{\textit{Baseline}}
& \ding{55} & \ding{55}           & 53.8 & 53.0 & 53.0 & 53.0 & 53.0 & 53.0 & 53.0 & 53.0 & 53.0 & 53.0 & 53.8 & 53.0 \\ 
\cmidrule{2-15} 
\multirow{2}{*}{\textit{SFT only}}
& \ding{55} & {\small \faSnowflake} & 73.1 & 68.5 & 71.1 & 70.0 & 68.5 & 67.7 & 65.5 & 58.6 & 64.9 & 60.9 & 66.9 & 66.1 \\
& \ding{55} & {\small \faFire}      & 72.6 & 68.8 & 70.6 & 70.0 & \textbf{68.6} & 67.9   & 65.1 & 60.0 & 65.2 & 61.6 & 67.0 & 66.2 \\
\cmidrule{2-15} 
\multirow{3}{*}{\textit{Align+SFT}}
& {\small \faSnowflake}   & {\small \faSnowflake}   & \textbf{73.6} & \textbf{69.6} & 70.9 & \textbf{70.8} & 67.8 & 67.8 & 65.6 & 61.0 & 65.1 & 62.2 & 67.5 & 66.7 \\
& {\small \faFire} &  {\small \faSnowflake}  & 73.7 & 69.2 & \textbf{71.3} & 70.7 & \textbf{68.0} & \textbf{68.0} & \textbf{66.1} & \textbf{62.7} & \textbf{66.6} & \textbf{62.9} & \textbf{68.0} & \textbf{67.2} \\
& {\small \faFire} & {\small \faFire} & 73.5 & 69.1 & 70.4 & 70.2 & 68.2 & 68.1 & 65.3 & 59.6 & 64.4 & 62.1 & 67.1 & 66.3 \\
\bottomrule
\end{tabular*}
}
\caption{
Performance of different training strategies on Qwen2.5-VL-7B across various languages. {\small \faSnowflake} and {\small \faFire} indicate that the vision encoder is frozen or unfrozen, respectively, during each stage. \ding{55} denotes that the stage is skipped. Bold values denote the best performance.
}
\label{tab:ve_training_strategies_full_x100}
\end{table*}

\subsection{Full Results of Fine-tuning on \textsc{PolyChartQA}-Train}
\label{app:more_exp_ft}

We provide the complete fine-tuning results of various multilingual LVLMs on \textsc{PolyChartQA}-Train. This extended analysis reports per-language accuracy across all ten languages, offering a detailed view of how fine-tuning impacts different linguistic settings and model scales. As shown in Table~\ref{tab:training_results_all_split}, all models exhibit consistent improvements after fine-tuning, with particularly large gains for smaller or previously weaker models. Results also show that fine-tuning yields the most significant relative improvements in low-resource languages such as Urdu, Bengali, and Hindi, where accuracies often increase by over 100\%, reflecting the strong transferability of multilingual chart instruction data. In contrast, high-resource languages such as English, Chinese, and French experience smaller yet consistent improvements, suggesting a saturation effect from stronger pretraining. Overall, these results indicate that fine-tuning primarily bridges multilingual reasoning gaps, especially in linguistically underrepresented settings.

\begin{table*}[t]
\centering
\setlength{\tabcolsep}{4pt}

\resizebox{\textwidth}{!}{%
\begin{tabular}{c|cccccc}
\toprule
\textbf{Model} & \textbf{EN} & \textbf{ZH} & \textbf{FR} & \textbf{ES} & \textbf{RU} & \textbf{JA} \\
\midrule
\textbf{Qwen2.5-VL-3B} & 67.4 & 59.6 & 61.8 & 62.5 & 58.0 & 48.8 \\
\hspace{1.5em}\textit{w/ fine-tuning} & 68.2 \textcolor{ForestGreen}{\scriptsize (+1.2\%)} & 64.1 \textcolor{ForestGreen}{\scriptsize (+7.6\%)} & 66.1 \textcolor{ForestGreen}{\scriptsize (+7.0\%)} & 65.4 \textcolor{ForestGreen}{\scriptsize (+4.6\%)} & 64.9 \textcolor{ForestGreen}{\scriptsize (+11.9\%)} & 63.1 \textcolor{ForestGreen}{\scriptsize (+29.3\%)} \\
\midrule
\textbf{Qwen2.5-VL-7B} & 60.5 & 58.3 & 57.2 & 59.0 & 56.8 & 55.6 \\
\hspace{1.5em}\textit{w/ fine-tuning} & 73.1 \textcolor{ForestGreen}{\scriptsize (+20.8\%)} & 68.5 \textcolor{ForestGreen}{\scriptsize (+17.5\%)} & 71.1 \textcolor{ForestGreen}{\scriptsize (+24.3\%)} & 70.0 \textcolor{ForestGreen}{\scriptsize (+18.6\%)} & 68.5 \textcolor{ForestGreen}{\scriptsize (+20.6\%)} & 67.7 \textcolor{ForestGreen}{\scriptsize (+21.8\%)} \\
\midrule
\textbf{InternVL-3-2B} & 43.7 & 35.3 & 30.8 & 33.5 & 25.6 & 26.9 \\
\hspace{1.5em}\textit{w/ fine-tuning} & 48.9 \textcolor{ForestGreen}{\scriptsize (+11.9\%)} & 46.5 \textcolor{ForestGreen}{\scriptsize (+31.7\%)} & 43.1 \textcolor{ForestGreen}{\scriptsize (+39.9\%)} & 41.6 \textcolor{ForestGreen}{\scriptsize (+24.2\%)} & 36.6 \textcolor{ForestGreen}{\scriptsize (+43.0\%)} & 39.4 \textcolor{ForestGreen}{\scriptsize (+46.5\%)} \\
\midrule
\textbf{InternVL-3-8B} & 54.1 & 39.4 & 43.4 & 45.8 & 38.1 & 39.7 \\
\hspace{1.5em}\textit{w/ fine-tuning} & 63.1 \textcolor{ForestGreen}{\scriptsize (+16.6\%)} & 57.3 \textcolor{ForestGreen}{\scriptsize (+45.4\%)} & 57.7 \textcolor{ForestGreen}{\scriptsize (+32.9\%)} & 58.0 \textcolor{ForestGreen}{\scriptsize (+26.6\%)} & 50.7 \textcolor{ForestGreen}{\scriptsize (+33.1\%)} & 53.1 \textcolor{ForestGreen}{\scriptsize (+33.8\%)} \\
\midrule
\textbf{PaliGemma2-3B} & 26.6 & 14.7 & 19.7 & 21.5 & 13.9 & 10.7 \\
\hspace{1.5em}\textit{w/ fine-tuning} & 33.9 \textcolor{ForestGreen}{\scriptsize (+27.4\%)} & 28.5 \textcolor{ForestGreen}{\scriptsize (+93.9\%)} & 32.3 \textcolor{ForestGreen}{\scriptsize (+64.0\%)} & 33.1 \textcolor{ForestGreen}{\scriptsize (+54.0\%)} & 30.0 \textcolor{ForestGreen}{\scriptsize (+115.8\%)} & 28.9 \textcolor{ForestGreen}{\scriptsize (+170.1\%)} \\
\midrule
\textbf{LLaVA-v1.6-7B} & 24.8 & 12.9 & 18.9 & 18.2 & 13.5 & 11.5 \\
\hspace{1.5em}\textit{w/ fine-tuning} & 36.6 \textcolor{ForestGreen}{\scriptsize (+47.6\%)} & 22.2 \textcolor{ForestGreen}{\scriptsize (+72.1\%)} & 33.6 \textcolor{ForestGreen}{\scriptsize (+77.8\%)} & 33.8 \textcolor{ForestGreen}{\scriptsize (+85.7\%)} & 24.6 \textcolor{ForestGreen}{\scriptsize (+82.2\%)} & 20.9 \textcolor{ForestGreen}{\scriptsize (+81.7\%)} \\
\bottomrule
\end{tabular}
}
\vspace{3pt}

\vspace{6pt}
\resizebox{\textwidth}{!}{%
\begin{tabular}{c|cccccc}
\toprule
\textbf{Model} & \textbf{AR} & \textbf{UR} & \textbf{HI} & \textbf{BN} & \textbf{\begin{tabular}{@{}c@{}}Avg.\\(w EN)\end{tabular}} & \textbf{\begin{tabular}{@{}c@{}}Avg.\\(w/o EN)\end{tabular}} \\
\midrule
\textbf{Qwen2.5-VL-3B} & 51.4 & 37.2 & 45.7 & 43.0 & 53.7 & 51.8 \\
\hspace{1.5em}\textit{w/ fine-tuning} & 59.0 \textcolor{ForestGreen}{\scriptsize (+14.8\%)} & 49.8 \textcolor{ForestGreen}{\scriptsize (+33.9\%)} & 56.8 \textcolor{ForestGreen}{\scriptsize (+24.3\%)} & 54.0 \textcolor{ForestGreen}{\scriptsize (+25.6\%)} & 61.1 \textcolor{ForestGreen}{\scriptsize (+13.8\%)} & 60.2 \textcolor{ForestGreen}{\scriptsize (+16.2\%)} \\
\midrule
\textbf{Qwen2.5-VL-7B} & 52.0 & 43.7 & 49.4 & 46.4 & 53.8 & 53.0 \\
\hspace{1.5em}\textit{w/ fine-tuning} & 65.5 \textcolor{ForestGreen}{\scriptsize (+26.0\%)} & 58.6 \textcolor{ForestGreen}{\scriptsize (+34.1\%)} & 64.9 \textcolor{ForestGreen}{\scriptsize (+31.4\%)} & 60.9 \textcolor{ForestGreen}{\scriptsize (+31.3\%)} & 66.9 \textcolor{ForestGreen}{\scriptsize (+24.3\%)} & 66.1 \textcolor{ForestGreen}{\scriptsize (+24.7\%)} \\
\midrule
\textbf{InternVL-3-2B} & 17.1 & 14.6 & 15.7 & 11.9 & 25.6 & 23.1 \\
\hspace{1.5em}\textit{w/ fine-tuning} & 21.6 \textcolor{ForestGreen}{\scriptsize (+26.3\%)} & 18.3 \textcolor{ForestGreen}{\scriptsize (+25.3\%)} & 20.7 \textcolor{ForestGreen}{\scriptsize (+31.8\%)} & 18.2 \textcolor{ForestGreen}{\scriptsize (+52.9\%)} & 33.3 \textcolor{ForestGreen}{\scriptsize (+30.1\%)} & 31.2 \textcolor{ForestGreen}{\scriptsize (+35.1\%)} \\
\midrule
\textbf{InternVL-3-8B} & 21.4 & 17.2 & 20.2 & 17.5 & 33.8 & 31.0 \\
\hspace{1.5em}\textit{w/ fine-tuning} & 26.6 \textcolor{ForestGreen}{\scriptsize (+24.3\%)} & 24.3 \textcolor{ForestGreen}{\scriptsize (+41.3\%)} & 26.4 \textcolor{ForestGreen}{\scriptsize (+30.7\%)} & 24.2 \textcolor{ForestGreen}{\scriptsize (+38.3\%)} & 44.0 \textcolor{ForestGreen}{\scriptsize (+30.2\%)} & 41.4 \textcolor{ForestGreen}{\scriptsize (+33.5\%)} \\
\midrule
\textbf{PaliGemma2-3B} & 15.9 & 12.2 & 14.3 & 10.2 & 16.3 & 14.9 \\
\hspace{1.5em}\textit{w/ fine-tuning} & 26.5 \textcolor{ForestGreen}{\scriptsize (+66.7\%)} & 26.2 \textcolor{ForestGreen}{\scriptsize (+114.8\%)} & 27.1 \textcolor{ForestGreen}{\scriptsize (+89.5\%)} & 22.7 \textcolor{ForestGreen}{\scriptsize (+122.5\%)} & 29.0 \textcolor{ForestGreen}{\scriptsize (+77.9\%)} & 28.4 \textcolor{ForestGreen}{\scriptsize (+90.6\%)} \\
\midrule
\textbf{LLaVA-v1.6-7B} & 12.0 & 7.7 & 10.0 & 6.7 & 13.9 & 12.4 \\
\hspace{1.5em}\textit{w/ fine-tuning} & 20.3 \textcolor{ForestGreen}{\scriptsize (+69.2\%)} & 20.2 \textcolor{ForestGreen}{\scriptsize (+162.3\%)} & 19.5 \textcolor{ForestGreen}{\scriptsize (+95.0\%)} & 19.2 \textcolor{ForestGreen}{\scriptsize (+186.6\%)} & 25.5 \textcolor{ForestGreen}{\scriptsize (+83.5\%)} & 24.0 \textcolor{ForestGreen}{\scriptsize (+93.5\%)} \\
\bottomrule
\end{tabular}
}
\vspace{3pt}
\caption{Fine-tuning Results using \textsc{PolyChartQA}-Train across different model families and sizes. Performance gains are highlighted in green.}
\label{tab:training_results_all_split}
\end{table*}


\subsection{Full Results of Ablation on Training Data Percentage}
\label{app:more_exp_abl_data_percentage}

The full results in Table~\ref{tab:full_training_data_percent} confirm a consistent positive correlation between data volume and model performance across all architectures.
The most substantial gains occur within the first 20–40\% of training data, after which improvements gradually plateau.
Notably, smaller models (e.g., InternVL3-2B) reach saturation earlier, while larger ones such as Qwen2.5-VL-7B continue to benefit steadily from additional data, underscoring their stronger data utilization capacity.

\begin{table*}[t]
\small
\centering
\setlength{\tabcolsep}{4pt} 
\begin{adjustbox}{max width=\textwidth}
\begin{tabular}{@{}l|c|cccccccccc|cc@{}}
\toprule
\textbf{Model} & \textbf{\% Data} & {\textbf{EN}} & {\textbf{ZH}} & {\textbf{FR}} & {\textbf{ES}} & {\textbf{RU}} & {\textbf{JA}} & {\textbf{AR}} & {\textbf{UR}} & {\textbf{HI}} & {\textbf{BN}} & {\textbf{Avg. (w EN)}} & {\textbf{Avg. (w/o EN)}} \\
\midrule
\multirow{6}{*}{\textbf{Qwen2.5-VL-3B}} & 0 & 67.4 & 59.6 & 61.8 & 62.5 & 58.0 & 48.8 & 51.4 & 37.2 & 45.7 & 43.0 & 53.7 & 51.8 \\
& 20 & 67.0 & 61.8 & 64.1 & 63.0 & 62.0 & 57.1 & 56.3 & 43.6 & 51.1 & 46.5 & 57.3 & 56.0 \\
& 40 & 67.5 & 62.6 & 65.4 & 64.5 & 63.3 & 60.4 & 57.5 & 46.5 & 53.8 & 50.4 & 59.2 & 58.1 \\
& 60 & \underline{68.4} & 63.8 & 66.1 & 65.4 & 64.9 & 62.1 & 58.7 & 49.0 & 56.6 & 53.5 & 60.8 & 59.8 \\
& 80 & \textbf{68.5} & \textbf{64.3} & \textbf{66.4} & \textbf{65.6} & 64.9 & \textbf{63.6} & \textbf{59.1} & \textbf{50.3} & \textbf{57.3} & \textbf{54.2} & \textbf{61.4} & \textbf{60.5} \\
& 100 & 68.2 & \underline{64.1} & \underline{66.1} & \underline{65.4} & \textbf{64.9} & 63.1 & \underline{59.0} & \underline{49.8} & \underline{56.8} & \underline{54.0} & \underline{61.1} & \underline{60.2} \\
\midrule
\multirow{6}{*}{\textbf{Qwen2.5-VL-7B}} & 0 & 60.5 & 58.3 & 57.2 & 59.0 & 56.8 & 55.6 & 52.0 & 43.7 & 49.4 & 46.4 & 53.8 & 53.0 \\
& 20 & 69.8 & 64.4 & 67.0 & 67.2 & 66.1 & 62.6 & 59.5 & 52.5 & 58.5 & 54.7 & 62.3 & 61.3 \\
& 40 & 71.9 & 67.2 & 69.7 & 69.2 & 67.6 & 65.6 & 61.9 & 55.7 & 61.7 & 57.3 & 64.8 & 63.9 \\
& 60 & \underline{72.2} & 66.9 & 69.7 & 69.0 & 67.7 & 65.6 & 61.6 & 55.4 & 61.5 & 57.5 & 64.8 & 63.8 \\
& 80 & 72.1 & 68.1 & \underline{70.1} & \underline{69.2} & \textbf{68.2} & 66.2 & 62.8 & 56.8 & 62.2 & 58.2 & 65.4 & 64.5 \\
& 100 & \textbf{73.1} & \textbf{68.5} & \textbf{71.1} & \textbf{70.0} & \underline{68.5} & \textbf{67.7} & \textbf{65.5} & \textbf{58.6} & \textbf{64.9} & \textbf{60.9} & \textbf{66.9} & \textbf{66.1} \\
\midrule
\multirow{6}{*}{\textbf{InternVL3-2B}} & 0 & 43.7 & 35.3 & 30.8 & 33.5 & 25.6 & 26.9 & 17.1 & 14.6 & 15.7 & 11.9 & 25.6 & 23.1 \\
& 20 & 47.3 & 41.6 & 38.9 & 39.5 & 32.5 & 33.9 & 19.4 & 18.1 & 19.5 & 17.0 & 30.7 & 28.5 \\
& 40 & 48.0 & 45.6 & 42.2 & 41.1 & 35.6 & 39.2 & 21.1 & 18.4 & 20.6 & 18.2 & 32.8 & 30.8 \\
& 60 & 50.0 & 46.9 & 44.4 & 43.0 & 37.3 & 40.4 & 22.6 & 19.2 & 21.2 & 19.0 & 34.2 & 32.1 \\
& 80 & \textbf{50.1} & \textbf{47.5} & \textbf{45.3} & \textbf{43.6} & \textbf{38.2} & \textbf{41.5} & \textbf{22.7} & \textbf{19.3} & \textbf{21.3} & \textbf{19.6} & \textbf{34.7} & \textbf{32.7} \\
& 100 & \underline{48.9} & \underline{46.5} & \underline{43.1} & \underline{41.6} & \underline{36.6} & \underline{39.4} & \underline{21.6} & \underline{18.3} & \underline{20.7} & \underline{18.2} & \underline{33.3} & \underline{31.2} \\
\midrule
\multirow{6}{*}{\textbf{InternVL3-8B}} & 0 & 54.1 & 39.4 & 43.4 & 45.8 & 38.1 & 39.7 & 21.4 & 17.2 & 20.2 & 17.5 & 33.8 & 31.0 \\
& 20 & 59.7 & 50.6 & 53.9 & 54.1 & 45.9 & 44.7 & 24.2 & 21.0 & 23.6 & 21.9 & 39.9 & 37.3 \\
& 40 & 61.7 & 55.6 & 56.9 & 56.5 & 49.0 & 49.4 & 26.6 & 23.1 & 24.8 & 23.7 & 42.6 & 40.1 \\
& 60 & 63.1 & 56.8 & 57.3 & 57.3 & 50.0 & 52.9 & 26.6 & 24.3 & 26.0 & 24.6 & 43.7 & 41.2 \\
& 80 & \textbf{63.7} & \textbf{58.3} & \textbf{58.2} & \textbf{58.4} & \textbf{51.3} & \textbf{54.7} & \textbf{27.3} & \textbf{25.6} & \textbf{26.9} & \textbf{24.7} & \textbf{44.7} & \textbf{42.2} \\
& 100 & \underline{63.1} & \underline{57.3} & \underline{57.7} & \underline{58.0} & \underline{50.7} & \underline{53.1} & \underline{26.6} & \underline{24.3} & \underline{26.4} & \underline{24.2} & \underline{44.0} & \underline{41.4} \\
\bottomrule
\end{tabular}
\end{adjustbox}
\caption{Overall performance on \textsc{PolyChartQA} benchmark across different fine-tuning data proportions. For each model category, the best score per column is in \textbf{bold} and the second-best is \underline{underlined}.}

\label{tab:full_training_data_percent}
\end{table*}

\section{Full Prompt Templates Used in Our Study}
\label{app:full_prompts}

In this section, we present all prompt templates used throughout our \textsc{PolyChartQA} data pipeline. This includes the pipeline prompts for data cleaning, generation, translation, and consistency checking.

\subsection{Prompts Used in Seed Data Preparation}
\label{app:prompt_seed_data}

\begin{figure*}[htbp]
  \centering
  \includegraphics[width=0.95\columnwidth]{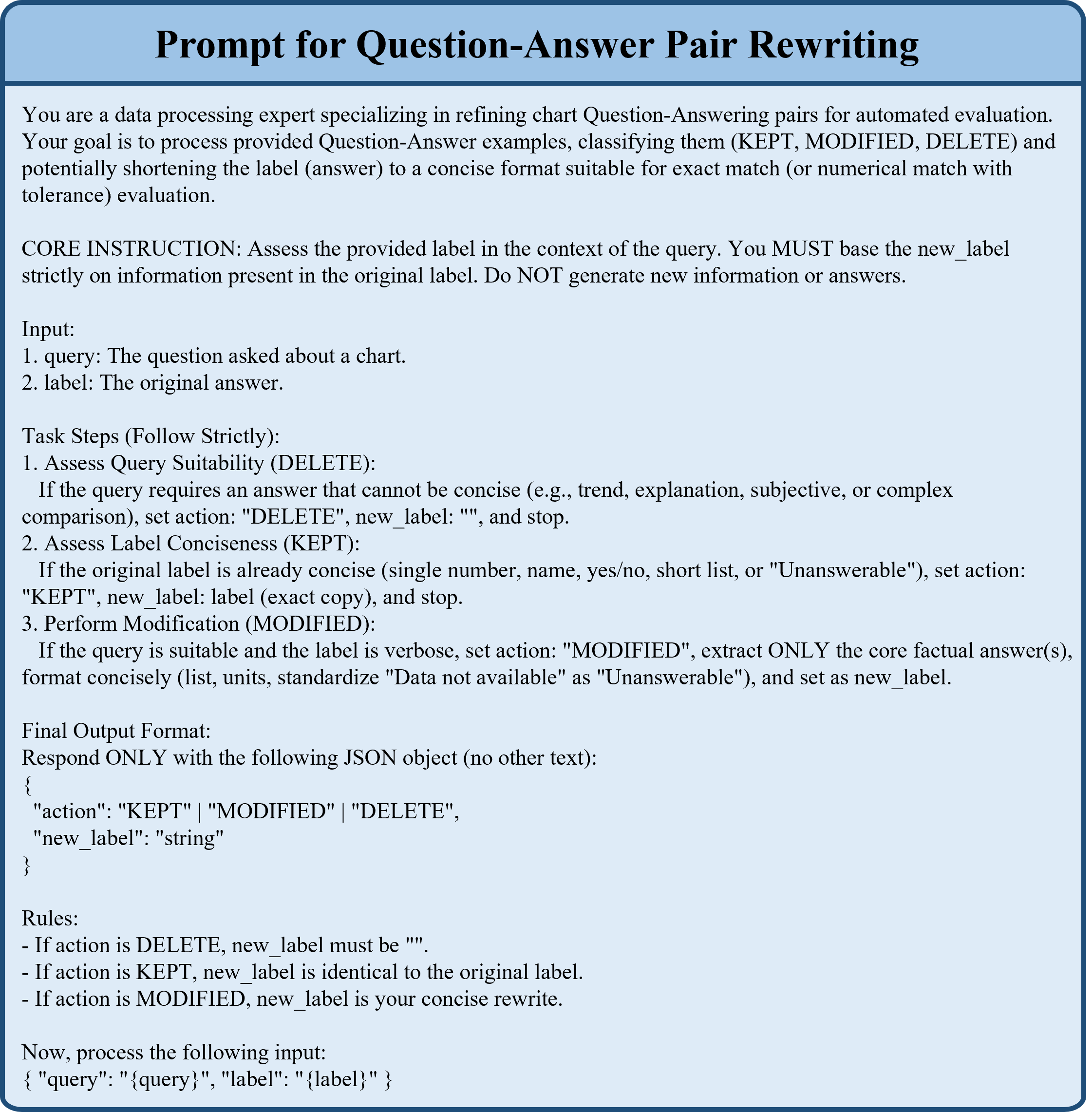}
  \caption{Prompt for question--answer pair rewriting.}
  \label{fig:prompt_rewrite}
\end{figure*}

\begin{figure*}[htbp]
  \centering
  \includegraphics[width=0.95\columnwidth]{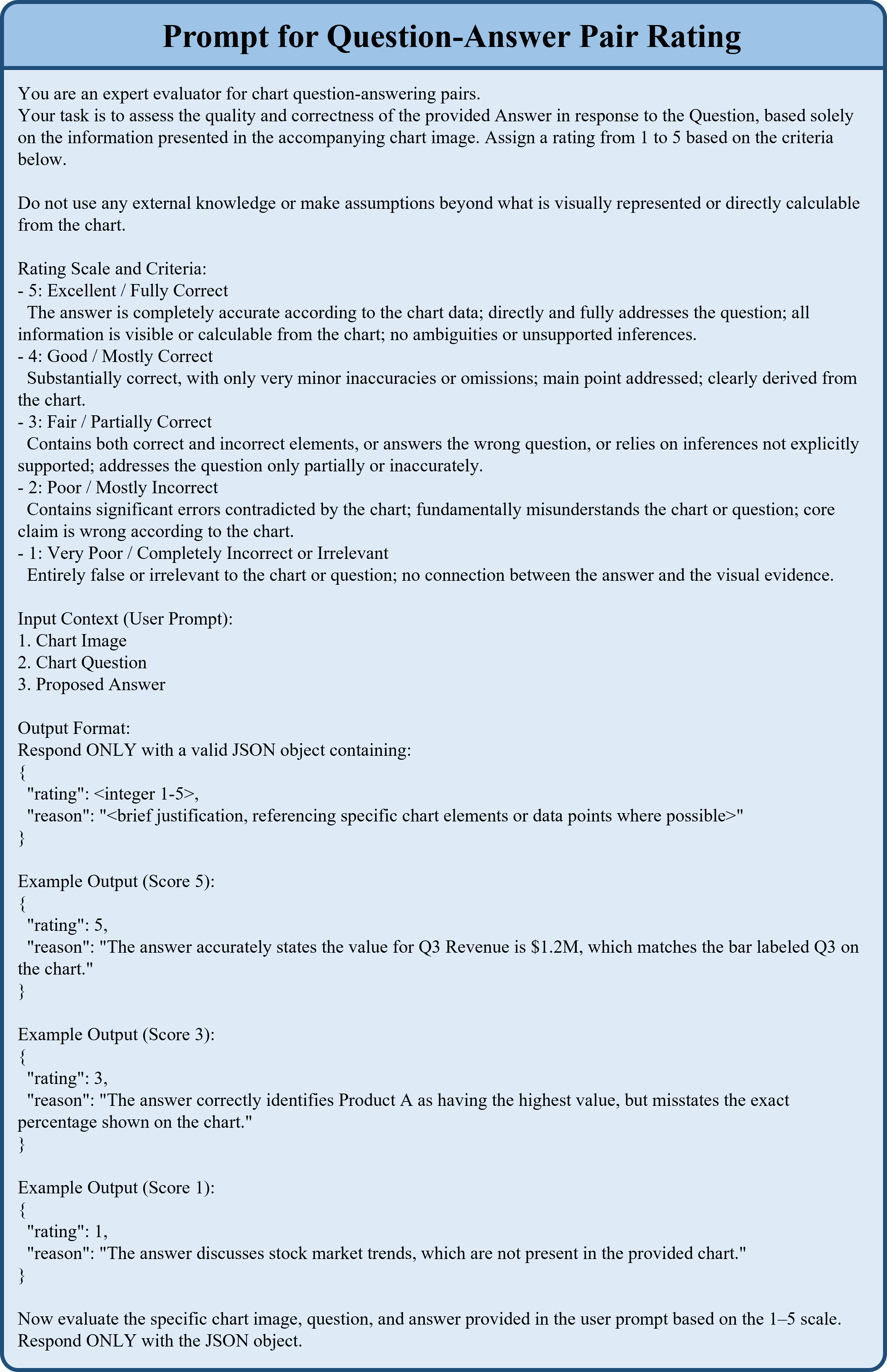}
  \caption{Prompt for question--answer pair rating.}
  \label{fig:prompt_rating}
\end{figure*}

\begin{figure*}[htbp]
  \centering
  \includegraphics[width=0.95\columnwidth]{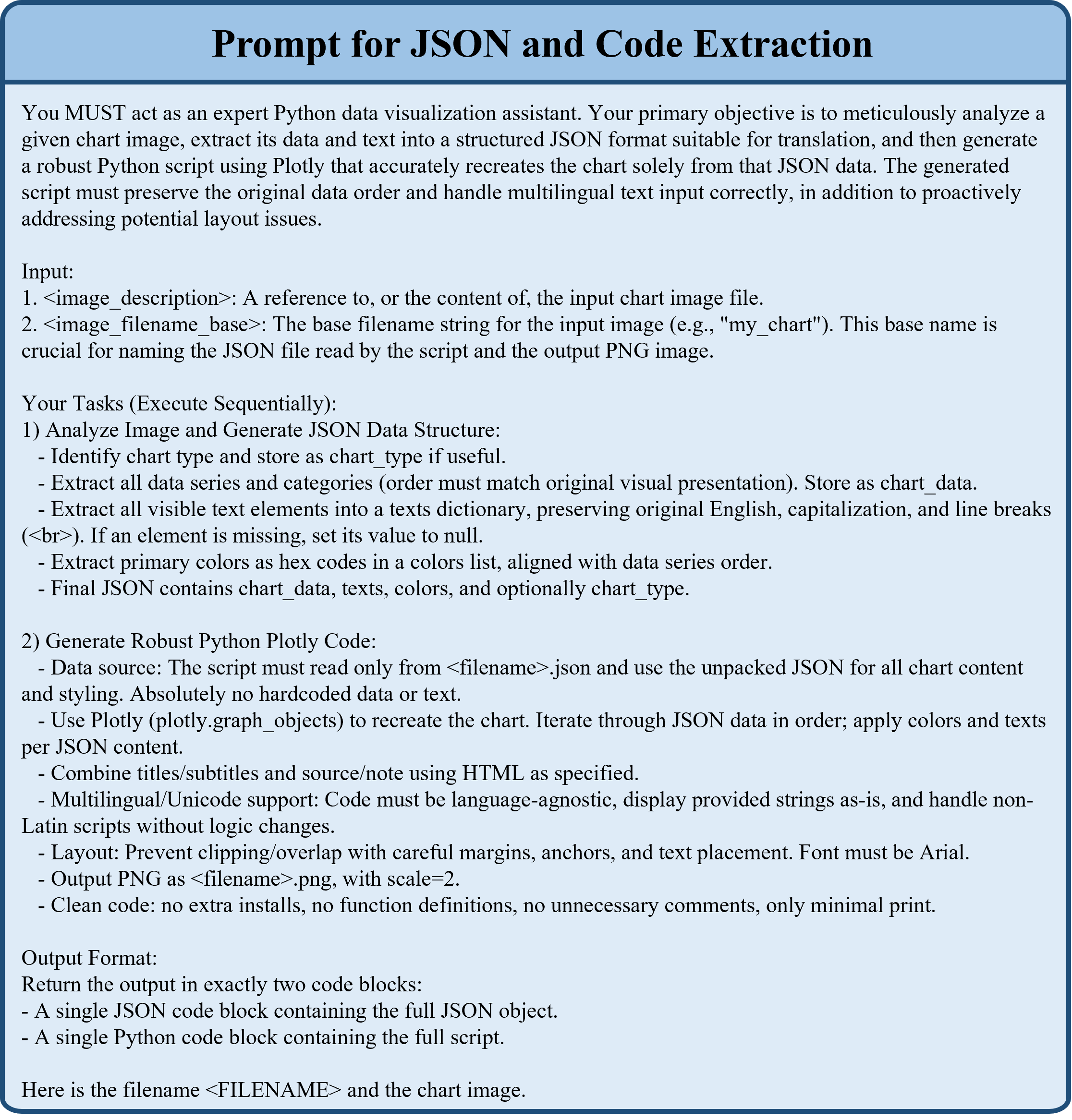}
  \caption{Prompt for JSON extraction and visualization code generation.}
  \label{fig:prompt_codegen}
\end{figure*}

\begin{figure*}[htbp]
  \centering
  \includegraphics[width=0.95\columnwidth]{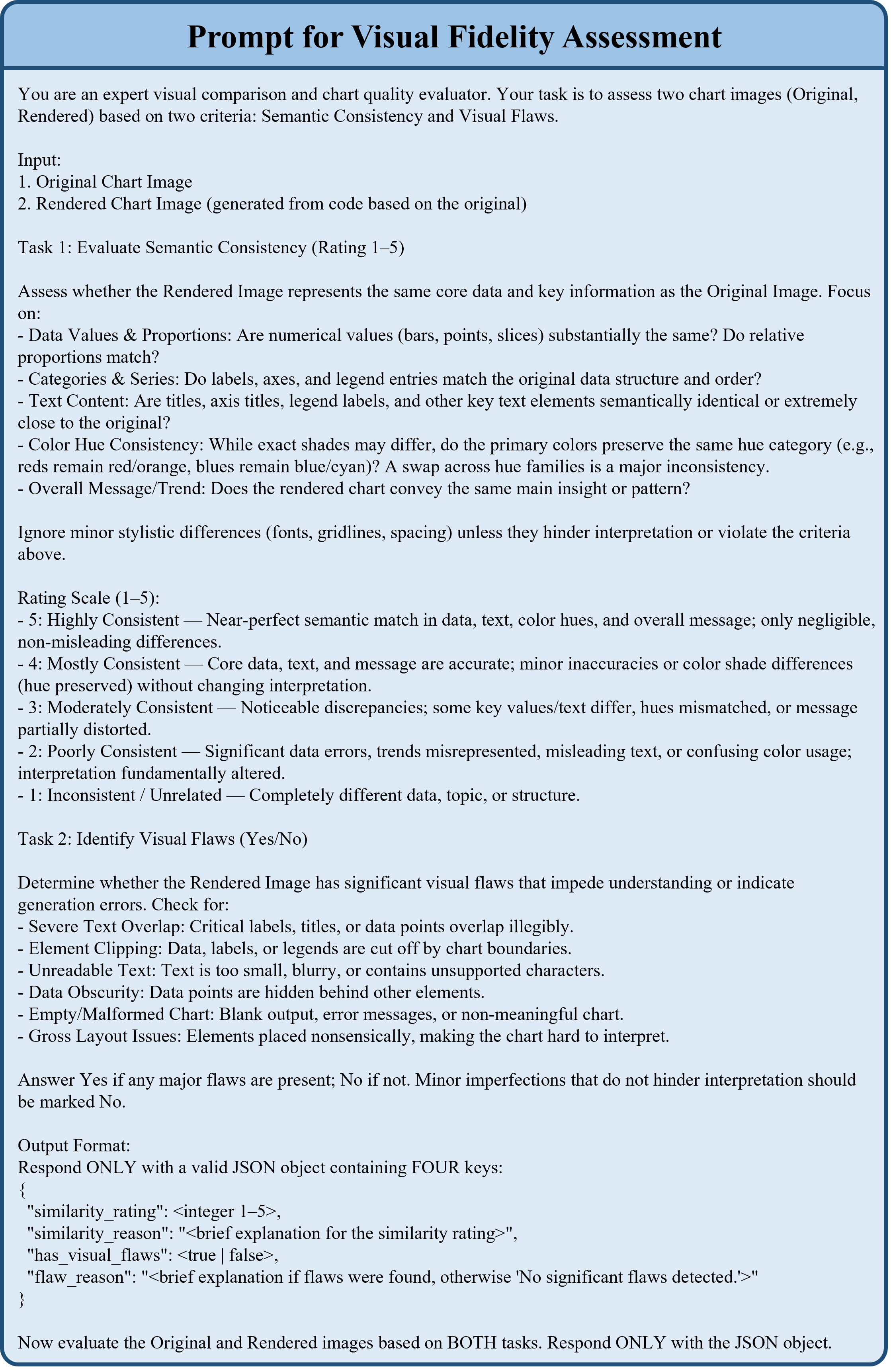}
  \caption{Prompt for visual fidelity checking.}
  \label{fig:prompt_visual}
\end{figure*}

\begin{figure*}[htbp]
  \centering
  \includegraphics[width=0.95\columnwidth]{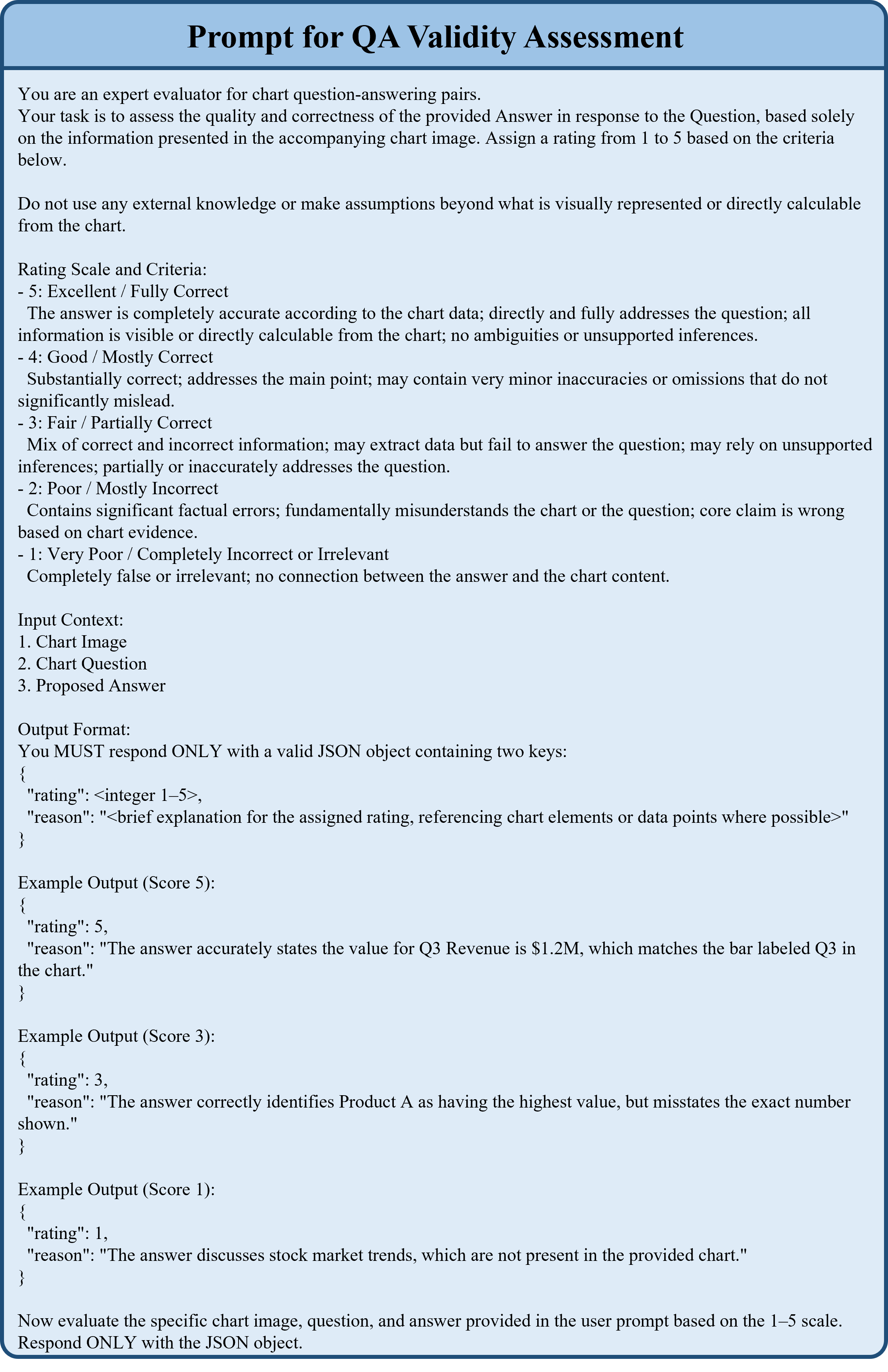}
  \caption{Prompt for QA validity checking.}
  \label{fig:prompt_qavalid}
\end{figure*}

The question-answer pair rewriting prompt used for \textbf{answer verification} of source datasets is shown in Figure~\ref{fig:prompt_rewrite}. The question-answer pair rating prompt used for \textbf{answer standardization} of source datasets is shown in Figure~\ref{fig:prompt_rating}. The prompt used for structured JSON extraction and visualization code generation during seed data construction is shown in Figure~\ref{fig:prompt_codegen}. The \textbf{visual fidelity} prompt used for quality control in seed data generation is shown in Figure~\ref{fig:prompt_visual}. The \textbf{QA validity} prompt used for quality control in seed data generation is shown in Figure~\ref{fig:prompt_qavalid}.

\subsection{Prompts Used in Multilingual Chart Generation}
\label{app:prompt_multilingual_data}

\begin{figure*}[htbp]
  \centering
  \includegraphics[width=0.95\columnwidth]{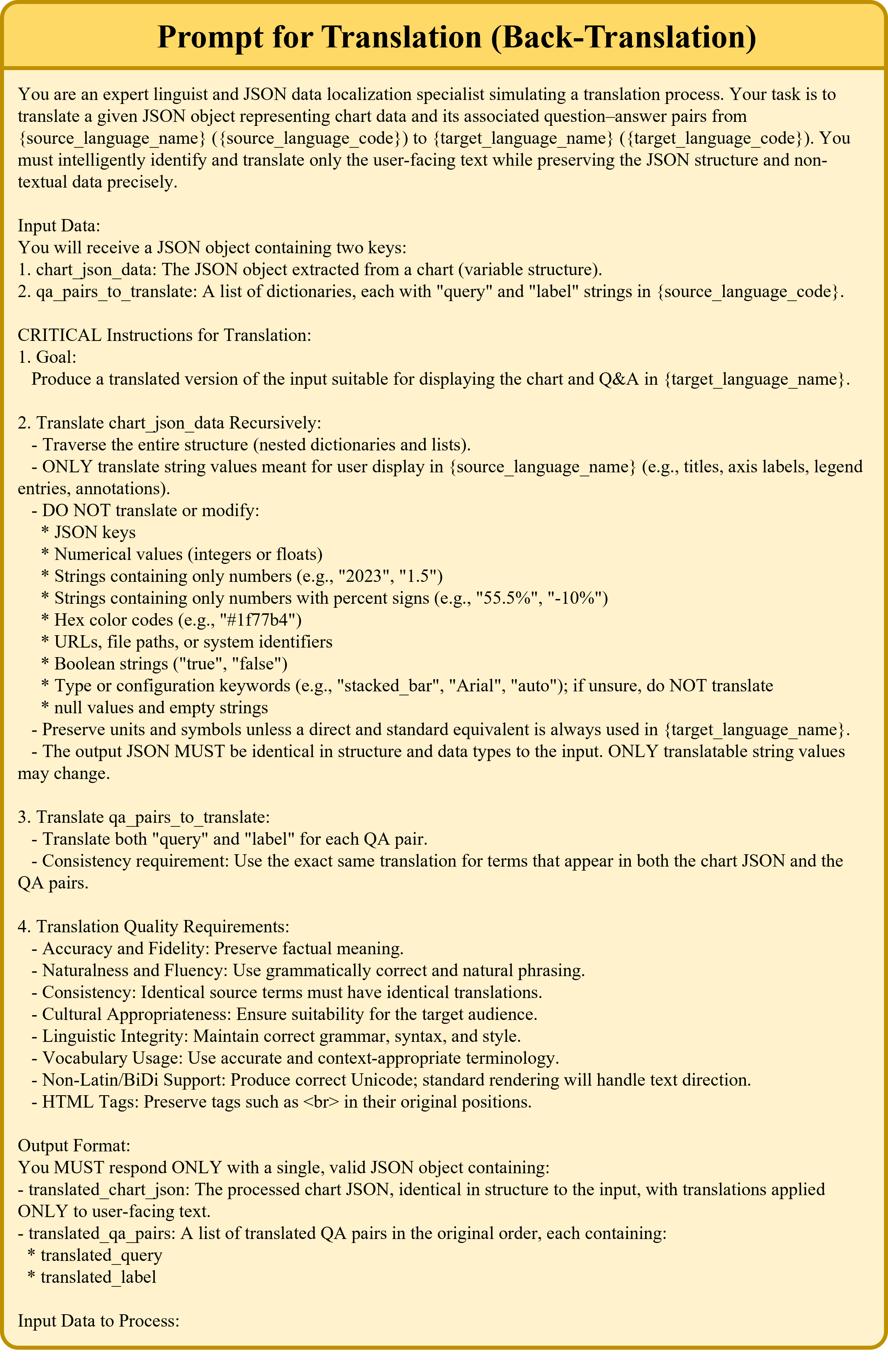}
  \caption{Prompt for multilingual translation.}
  \label{fig:prompt_translation}
\end{figure*}

\begin{figure*}[htbp]
  \centering
  \includegraphics[width=0.95\columnwidth]{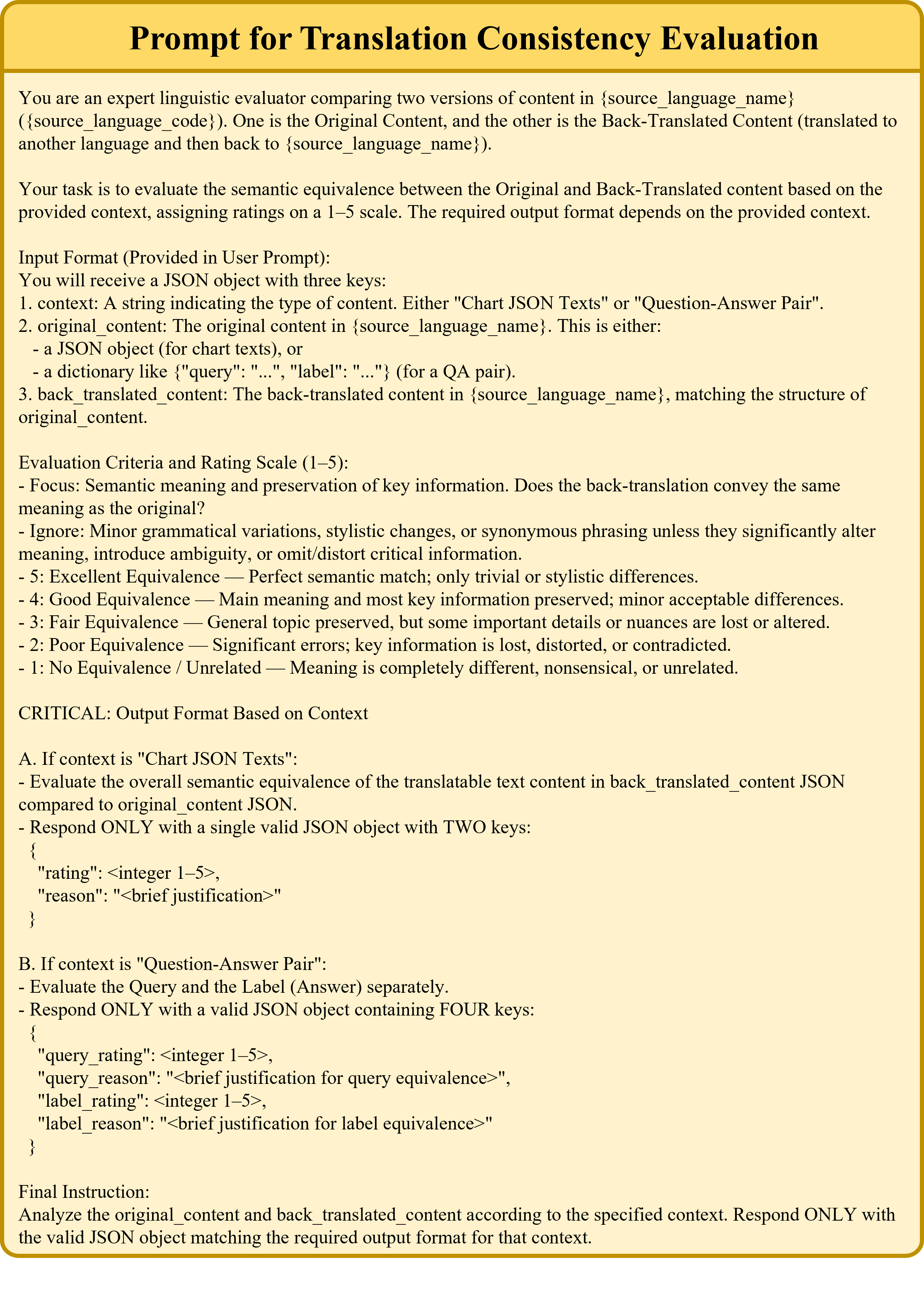}
  \caption{Prompt for translation consistency.}
  \label{fig:prompt_translation_consistency}
\end{figure*}

The \textbf{translation} prompt used for multilingual text translation is shown in Figure~\ref{fig:prompt_translation}. The \textbf{translation consistency} prompt used for back-translation verification is shown in Figure~\ref{fig:prompt_translation_consistency}.

\onecolumn

\end{document}